\documentclass[10pt,twocolumn,letterpaper]{article}
\usepackage[table]{xcolor} 

\usepackage[pagenumbers]{iccv} 

\usepackage{comment}
\usepackage{amsmath}
\usepackage{amssymb}
\usepackage{enumerate}

\usepackage{graphicx}
\usepackage{amsthm}           
\usepackage{booktabs}

\usepackage{epsfig}
\usepackage{tabularx}
\usepackage{booktabs}
\usepackage{relsize}
\usepackage{multirow}
\usepackage{scrextend}           
\usepackage{array}               
\usepackage{makecell}            
\usepackage{subcaption}          
\usepackage{wasysym}             
\usepackage{lipsum}              
\usepackage{esvect}              
\usepackage{soul}                
\usepackage{float}
\usepackage{caption}
\usepackage{bm}
\usepackage{enumitem}            
\usepackage{empheq}
\usepackage{gensymb}             
\usepackage{wrapfig}             
\usepackage{xspace}              
\usepackage[thicklines]{cancel}  
\usepackage{arydshln}            
\usepackage{pifont}              
\usepackage{hhline}              
\usepackage{tocloft}             

\setlist[itemize, 1]{label =\raisebox{-0.4\height}{\scalebox{1.7}{\textbullet}}}
\setlist[itemize]{noitemsep, topsep=0cm, leftmargin=4mm}

\allowdisplaybreaks              

\graphicspath{{images/}}
\sethlcolor{ood_color}

\advance\cftsecnumwidth 0.25cm\relax
\advance\cftsubsecnumwidth 0.08cm\relax
\advance\cftsubsecnumwidth 0.15cm\relax

\newtheorem{theorem}{Theorem}

\newtheorem{lemma}{Lemma}

\makeatletter
\@namedef{ver@everyshi.sty}{}
\makeatother
\usepackage{tikz} 
\usetikzlibrary{shapes.geometric} 
\usetikzlibrary{calc}  
\usetikzlibrary{arrows}
\usetikzlibrary{arrows.meta}
\usetikzlibrary{shapes.arrows}
\usetikzlibrary{fadings, shadows}
\usetikzlibrary{decorations.text}
\def\centerarc[#1](#2)(#3:#4:#5)
    { \draw[#1] ($(#2)+({#5*cos(#3)},{#5*sin(#3)})$) arc (#3:#4:#5); }

\usepackage[ruled,linesnumbered]{algorithm2e}  
\SetKwIF{If}{ElseIf}{Else}{if}{}{else if}{else}{}
\SetKwFor{For}{for}{do}{end}
\SetKwComment{Comment}{$\triangleright$\ }{}

\SetCommentSty{mycommfont}

\definecolor{my_blue}{rgb}{0.2, 0.6, 1}  
\definecolor{my_magenta}{rgb}{1.0, 0.2, 0.6} 
\definecolor{my_yellow}{rgb}{1.0, 0.8, 0.2} 
\definecolor{my_green}{rgb}{0.0, 0.9, 0.24}
\definecolor{my_green_2}{rgb}{0.0, 0.4, 0.0}

\definecolor{sns_blue}{rgb}{0.21, 0.06, 0.42}    
\definecolor{sns_violet}{rgb}{0.45, 0.12, 0.51}  
\definecolor{sns_orange}{rgb}{0.75, 0.22, 0.46}  
\definecolor{sns_yellow}{rgb}{0.99, 0.91, 0.66}  

\definecolor{white}{rgb}{1.0, 1.0, 1.0}
\definecolor{darkGreen}{rgb}{0.01, 0.8, 0.24}
\definecolor{darkGreen2}{rgb}{0.22,0.42, 0.33}
\definecolor{darkGreen3}{rgb}{0.20,0.66, 0.33}
\definecolor{cvprblue}{rgb}{0.21,0.49,0.74}
\definecolor{LightCyan}{rgb}{0.88,1,1}
\definecolor{lightgreen}{HTML}{90EE90}
\definecolor{new_green}{rgb}{0.75,0.97,0.44}
\definecolor{Gray}{gray}{0.95}
\definecolor{lightgray}{rgb}{0.96, 0.96, 0.96}

\definecolor{set1_cyan}{rgb}{0.23, 0.87, 1.0}
\definecolor{building}{rgb}{0.2, 0.33, 0.33}

\definecolor{my_violet}{rgb}{0.79, 0.40, 1} 
\definecolor{my_yellow_2}{rgb}{0.9, 0.8, 0.54}
\definecolor{my_red}{rgb}{1,0,0}
\definecolor{my_purple}{rgb}{0.27,0.8, 0.8}
\definecolor{my_orange}{rgb}{1.0,0.6,0.35}
\definecolor{my_golden}{rgb}{1.0, 0.75, 0.0}
\colorlet{my_gray}{gray!12}

\definecolor{projectionColor}{rgb}{0.2, 0.6, 1}
\definecolor{rayColor}{rgb}{0.0,0.0,0.0}
\definecolor{axisColor}{rgb}{0.0, 0.0, 0.0}
\colorlet{projectionBorderShade}{rayColor!100}
\colorlet{projectionFillShade}{projectionColor!20}
\colorlet{rayShade}{my_yellow}
\colorlet{axisShade}{axisColor!20}
\colorlet{axisShadeDark}{axisColor!100}

\definecolor{backward_color}{rgb}{1.0, 0.6, 0.2}
\definecolor{forward_color}{rgb}{0.2, 1.0, 0.6}

\definecolor{gain}{HTML}{34a853}
\definecolor{lost}{HTML}{ea4335}

\colorlet{proposedShade}{darkGreen}
\colorlet{vanillaShade}{red!90}

\colorlet{theme_color}{sns_orange}
\colorlet{theme_color_light}{sns_yellow!25}

\colorlet{ood_color}{gray!20}
\colorlet{methodColor}{cyan!20}

\newcommand{\noIndentHeading}[1]{\noindent\textbf{#1}}

\definecolor{XLcolor}{rgb}{0.858, 0.188, 0.478}

\newcommand{\thatIs}{\textit{i.e.}\xspace}

\newcommand{\cmark}{\checkmark}

\newcommand{\identity}{{\bf{I}}}

\newcommand{\scaleFraction}{0.9}

\newcommand{\myTopRule}{\Xhline{2\arrayrulewidth}}

\newcolumntype{t}{!{\vrule width 1.5\arrayrulewidth}}
\newcolumntype{m}{!{\vrule width 2.5\arrayrulewidth}}
\newcolumntype{a}{>{\columncolor{theme_color_light}}l}
\newcolumntype{b}{>{\columncolor{theme_color_light}}c}
\newcolumntype{d}{>{\columncolor{ood_color}}c}

\colorlet{cyan_highlight}{my_blue!85}

\colorlet{darkGreen_highlight}{darkGreen!75}

\colorlet{my_magenta_highlight}{my_magenta!50}

\colorlet{my_yellow_highlight}{my_yellow!55}

\providecommand\rightarrowRHD{\relbar\joinrel\mathrel\RHD}

\newcommand{\uparrowRHD}  {\rotatebox[origin=c]{90}{$\rightarrowRHD$}}
\newcommand{\downarrowRHD}{\rotatebox[origin=c]{270}{$\rightarrowRHD$}}

\newcommand{\uparrowRHDSmall}  {\raisebox{0.05\normalbaselineskip}{\scalebox{0.7}{\uparrowRHD}}}   
\newcommand{\downarrowRHDSmall}{\raisebox{0.07\normalbaselineskip}{\scalebox{0.7}{\downarrowRHD}}} 
\newcommand{\rightarrowRHDSmall}{\raisebox{0.05\normalbaselineskip}{\scalebox{0.75}{$\rightarrowRHD$}}}   

\newcommand{\monoThreeD}{Mono3D\xspace}

\newcommand{\twoD}{$2$D\xspace}
\newcommand{\threeD}{$3$D\xspace}

\newcommand{\iouTwoD}{IoU$_{2\text{D}}$\xspace}
\newcommand{\iouThreeD}{IoU$_{3\text{D}}$\xspace}

\newcommand{\lidar}{LiDAR\xspace}

\newcommand{\radar}{radar\xspace}

\newcommand{\bev}{BEV\xspace}

\newcommand{\resNetEighteen}{ResNet-18\xspace}

\newcommand{\vit}{ViT\xspace}

\newcommand{\kitti}{KITTI\xspace}
\newcommand{\nuscenes}{nuScenes\xspace}

\newcommand{\val}{Val\xspace}

\newcommand{\carla}{CARLA\xspace}

\newcommand{\coda}{CODa\xspace}

\newcommand{\ap}{AP\xspace}
\newcommand{\apMath}{\text{\ap}}
\newcommand{\apThreeD}{$\apMath_{3\text{D}}$\xspace}

\newcommand{\MDE}{MDE\xspace}

\newcommand{\bracketPercentage}{[\%]}

\newcommand{\apThreeDSeventy}{\ap$_{\!3\text{D}\!}$ 70\xspace}
\newcommand{\apThreeDFifty}{\ap$_{\!3\text{D}\!}$ 50\xspace}
\newcommand{\apThreeDThirty}{\ap$_{\!3\text{D}\!}$ 30\xspace}

\newcommand{\first}[1]{$\textcolor{sns_blue}{\mathbf{#1}}$}
\newcommand{\second}[1]{$\textcolor{sns_orange}{\mathbf{#1}}$}
\newcommand{\firstKey}[1]{\textcolor{sns_blue}{\textbf{#1}}}
\newcommand{\secondKey}[1]{\textcolor{sns_orange}{\textbf{#1}}}
\newcommand{\sota}{SoTA\xspace}

\newcommand{\mathDash}{$-$}

\newcommand{\gupNet}{GUP Net\xspace}
\newcommand{\deviant}{DEVIANT\xspace}

\newcommand{\pseudoLidar}{Pseudo-{\lidar}\xspace}

\newcommand{\monodetr}{MonoDETR\xspace}

\newcommand{\bevHeight}{BEVHeight\xspace}

\newcommand{\ood}{OOD\xspace}
\newcommand{\outDomain}{OOD\xspace}
\newcommand{\inDomain}{ID\xspace}
\newcommand{\camConvs}{CAM-Convs\xspace}
\newcommand{\Variation}{Height\xspace}
\newcommand{\variation}{height\xspace}
\newcommand{\variations}{heights\xspace}
\newcommand{\zeroThreeD}{\bm{0}}
\newcommand{\egoHeight}{H}
\newcommand{\egoHeightChange}{\Delta H}
\newcommand{\uniDrive}{UniDrive\xspace}
\newcommand{\uniDrivePlus}{UniDrive++\xspace}

\newcommand{\newDepth}{\depthPred_{\egoHeightChange}}
\newcommand{\oldDepth}{\depthPred_{0}}

\newcommand{\newDepthGround}{{}^{g}\!\newDepth}
\newcommand{\oldDepthGround}{{}^{g}\!\oldDepth}
\newcommand{\newDepthRegress}{{}^{r}\!\newDepth}
\newcommand{\oldDepthRegress}{{}^{r}\!\oldDepth}
\newcommand{\depthSlope}{\beta}
\newcommand{\depthGTMax}{\depthGT_{max}}
\newcommand{\depthGTMin}{\depthGT_{min}}
\newcommand{\heightImage}{h}
\newcommand{\pixUBottom}{\pixU_b}
\newcommand{\pixVBottom}{\pixV_b}
\newcommand{\pixUCenter}{\pixU_c}
\newcommand{\pixVCenter}{\pixV_c}
\newcommand{\heightBoxTwoD}{h_{2D}}
\newcommand{\pixUCenterTwoD}{\pixU_{c,2D}}
\newcommand{\pixVCenterTwoD}{\pixV_{c,2D}}
\newcommand{\relu}{ReLU\xspace}
\newcommand{\plucker}{Plucker\xspace}
\newcommand{\ray}{\overrightarrow{r}}
\newcommand{\trend}{trend\xspace}
\newcommand{\trends}{trends\xspace}

\newcommand{\depthPred}{\hat{\posZ}}

\newcommand{\depthGT}{\posZ}

\newcommand{\normalVar}{\sigma^2}

\newcommand{\expect}{\mathbb{E}}

\newcommand{\noise}{\eta}

\newcommand{\intrinsic}{\mathbf{K}}

\newcommand{\extrinsicTrans}{\mathbf{T}}

\newcommand{\realDomain}{\mathbb{R}}

\newcommand{\varX}{X}
\newcommand{\varY}{Y}
\newcommand{\varZ}{Z}

\newcommand{\posZ}{z}

\newcommand{\rotation}{\mathbf{R}}
\newcommand{\translation}{\mathbf{t}}

\newcommand{\pixU}{u}
\newcommand{\pixV}{v}

\newcommand{\ppointU}{u_0} 
\newcommand{\ppointV}{v_0} 
\newcommand{\focal}{f}

\newcommand{\methodName}{CHARM3R\xspace}
\newcommand{\methodNameFull}{Camera Height Robust Monocular 3D Detector\xspace}

\newcommand{\paperTitle}{\textcolor{theme_color}{\methodName}: Towards Unseen \textcolor{theme_color}{C}amera \textcolor{theme_color}{H}eight \textcolor{theme_color}{R}obust \textcolor{theme_color}{M}onocular \textcolor{theme_color}{3}D Detecto\textcolor{theme_color}{r}\xspace}

\definecolor{iccvblue}{rgb}{0.21,0.49,0.74}
\usepackage[pagebackref,breaklinks,colorlinks,allcolors=iccvblue]{hyperref}

\def\paperID{7859} 
\def\confName{ICCV}
\def\confYear{2025}

\title{\paperTitle}

\author{Abhinav Kumar$^{1}$\quad Yuliang Guo$^{2}$\quad Zhihao Zhang$^{1}$\quad Xinyu Huang$^{2}$\quad Liu Ren$^{2}$\quad Xiaoming Liu$^{1}$\\
~~~~~$^{1}$Michigan State University\qquad~~$^{2}$Bosch Research North America, Bosch Center for AI~\\
{\small\tt~~$^{1}$[kumarab6,zhan2365,liuxm]@msu.edu ~~~ \tt\small$^{2}$[yuliang.guo2,xinyu.huang,liu.ren]@us.bosch.com}\\
{\small\url{https://github.com/abhi1kumar/CHARM3R}}
}

\begin{document}
\maketitle

\begin{abstract}
    Monocular 3D object detectors, while effective on data from one ego camera height, struggle with unseen or out-of-distribution camera heights.
    Existing methods often rely on Plucker embeddings, image transformations or data augmentation.
    This paper takes a step towards this understudied problem by first investigating the impact of camera height variations on state-of-the-art (SoTA) Mono3D models.
    With a systematic analysis on the extended CARLA dataset with multiple camera heights, we observe that depth estimation is a primary factor influencing performance under height variations.
    We mathematically prove and also empirically observe consistent negative and positive \trends in mean depth error of regressed and ground-based depth models, respectively, under camera height changes.
    To mitigate this, we propose Camera Height Robust Monocular 3D Detector (CHARM3R), which averages both depth estimates within the model.
    CHARM3R improves generalization to unseen camera heights by more than $45\%$, achieving \sota performance on the CARLA dataset.
\end{abstract}

\addtocontents{toc}{\protect\setcounter{tocdepth}{-2}}
\section{Introduction}\label{sec:intro}

    Monocular \threeD object detection (\monoThreeD) task uses a single image to determine both the \threeD location and dimensions of objects.
    This technology is essential for augmented reality \cite{alhaija2018augmented,Xiang2018RSS,park2019pix,merrill2022symmetry}, robotics \cite{saxena2008robotic}, and self-driving cars \cite{park2021pseudo,kumar2022deviant,li2022bevformer}, where accurate \threeD understanding of the environment is crucial.
    Our research specifically focuses on using \threeD object detectors applied to autonomous vehicles (AVs), as they have unique challenges and requirements.

    AVs necessitate detectors that are robust to a wide range of intrinsic and extrinsic factors, including intrinsics \cite{brazil2023omni3d}, domains \cite{li2024unimode}, object size \cite{kumar2024seabird}, rotations \cite{zhou2021monoef,moon2023rotation}, weather conditions \cite{lin2024monotta,oh2024monowad}, and adversarial examples \cite{zhu2023understanding}.
    Existing research primarily focusses on generalizing object detectors to these failure modes.
    However, this work investigates the generalization of \monoThreeD to another type, which, thus far, has been relatively understudied in the literature – {\it \monoThreeD generalization to unseen ego camera \variations}.

        \begin{figure}[!t]
            \centering
            \includegraphics[width=0.97\linewidth]{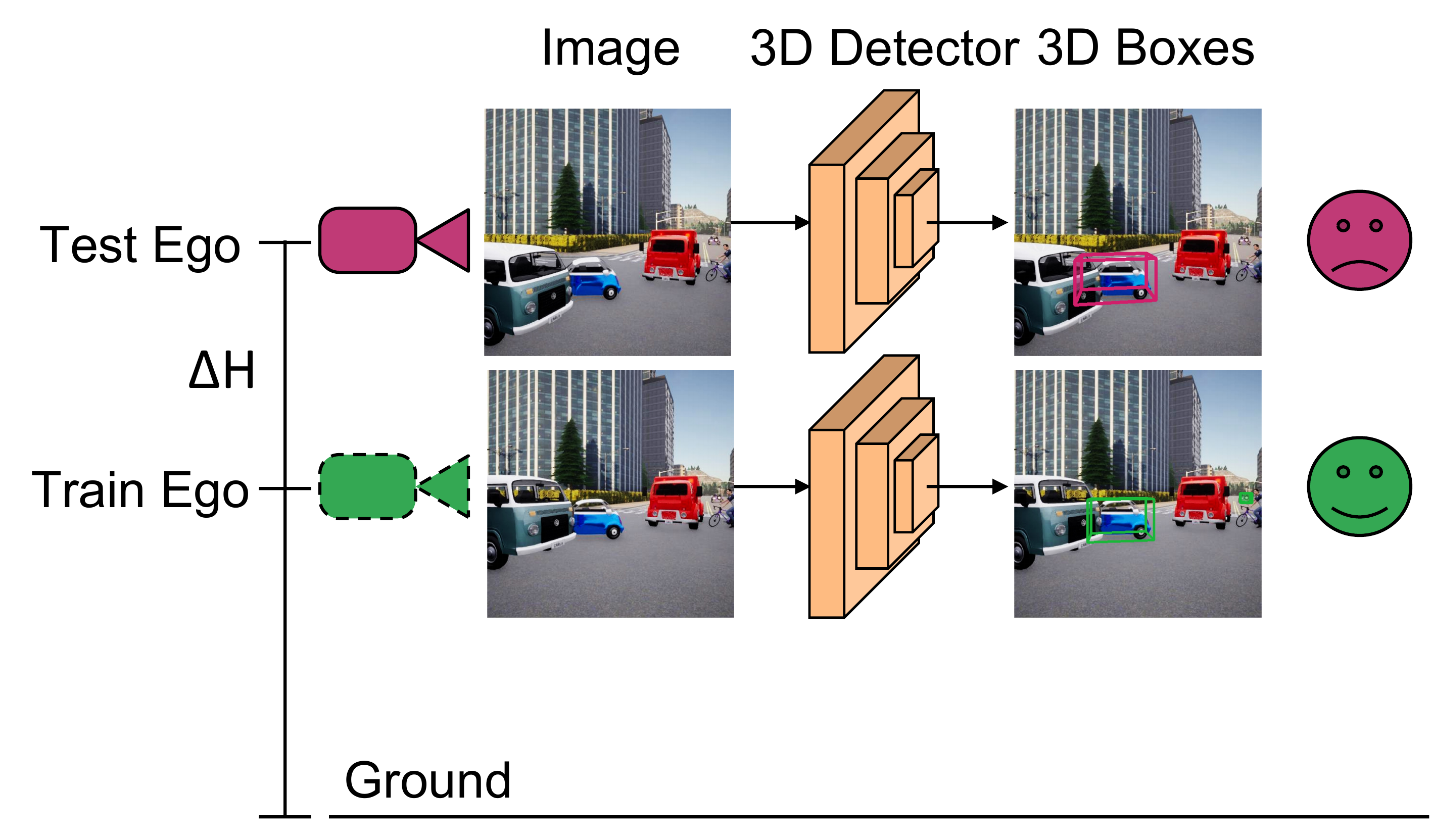}
            \vspace{-1mm}
            \caption{\textbf{Teaser.} Changing ego \variation at inference quickly \textbf{drops} \monoThreeD performance of \sota detectors. A \variation change $\egoHeightChange$ of $0.76m$ in inference drops \apThreeD \bracketPercentage~by absolute $35$ points.
            \vspace{-3mm} }
            \label{fig:teaser}
        \end{figure}

        \begin{figure*}[!t]
            \centering
            \begin{subfigure}{.32\linewidth}
                \includegraphics[width=\linewidth]{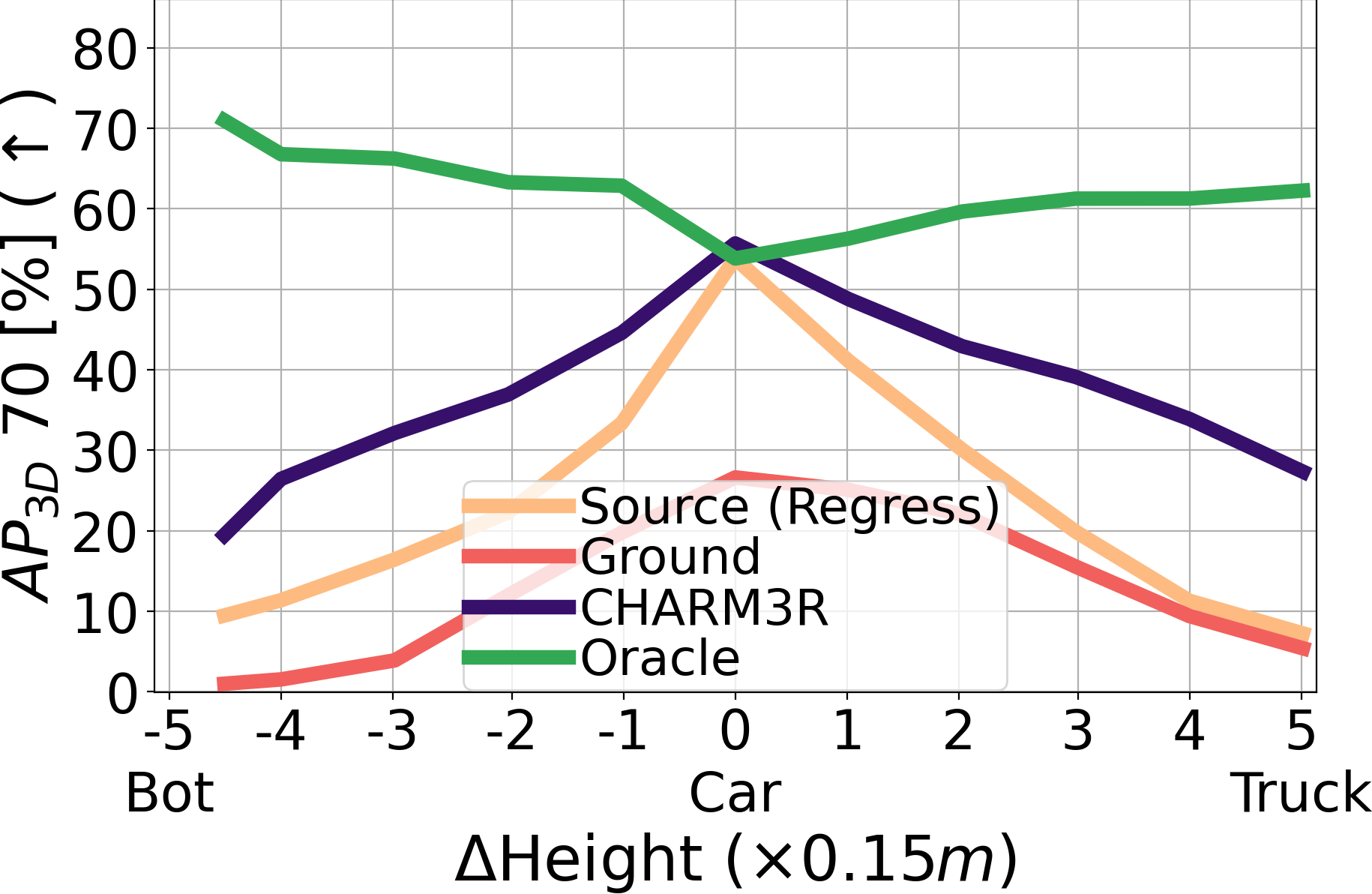}
                \caption{\apThreeDSeventy \bracketPercentage{} Results.}
                \label{fig:reason_seventy}
            \end{subfigure}%
            \hfill
            \begin{subfigure}{.32\linewidth}
                \includegraphics[width=\linewidth]{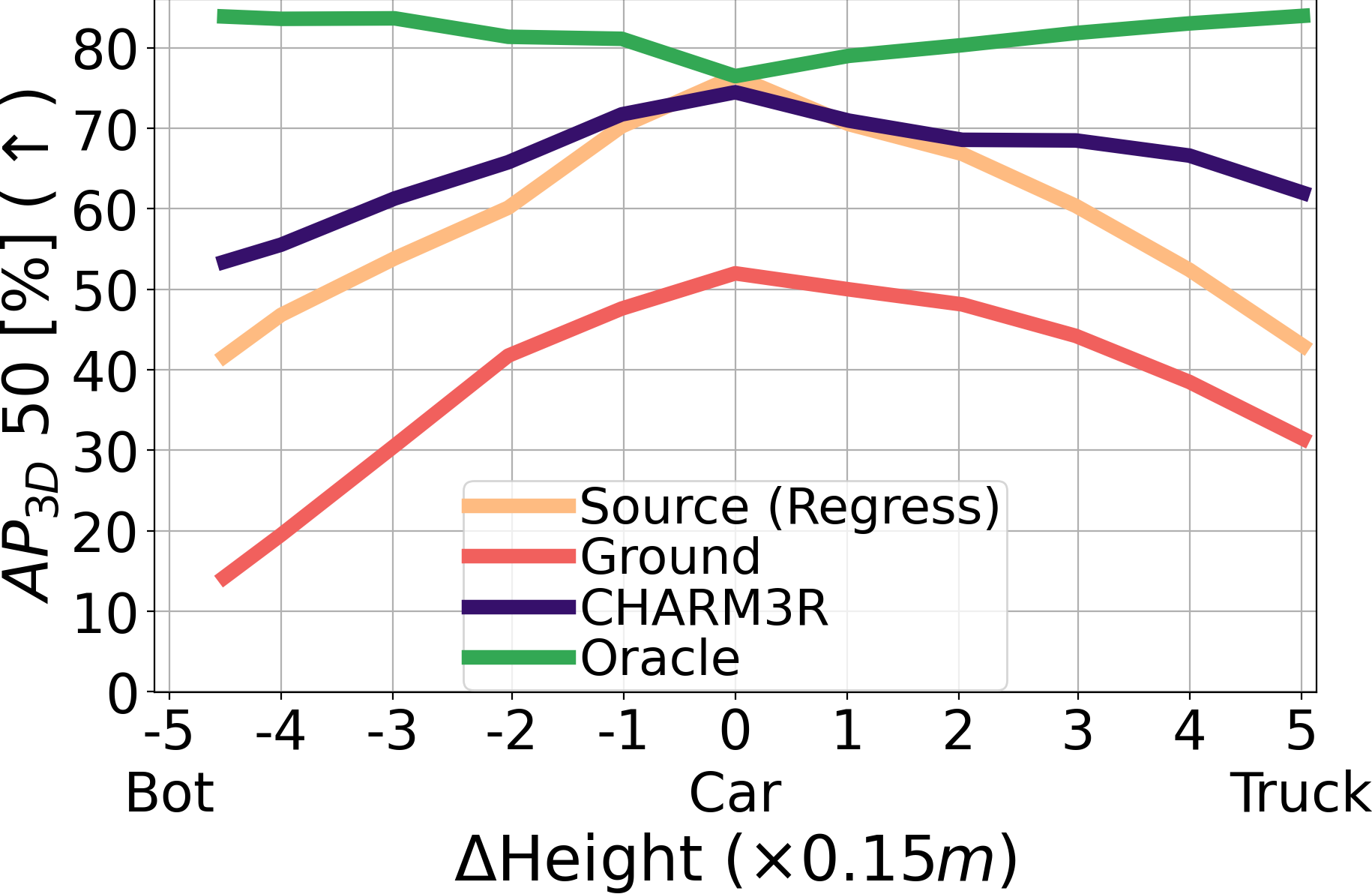}
                \caption{\apThreeDFifty \bracketPercentage{} Results.}
                \label{fig:reason_fifty}
            \end{subfigure}
            \hfill
            \begin{subfigure}{.32\linewidth}
                \includegraphics[width=\linewidth]{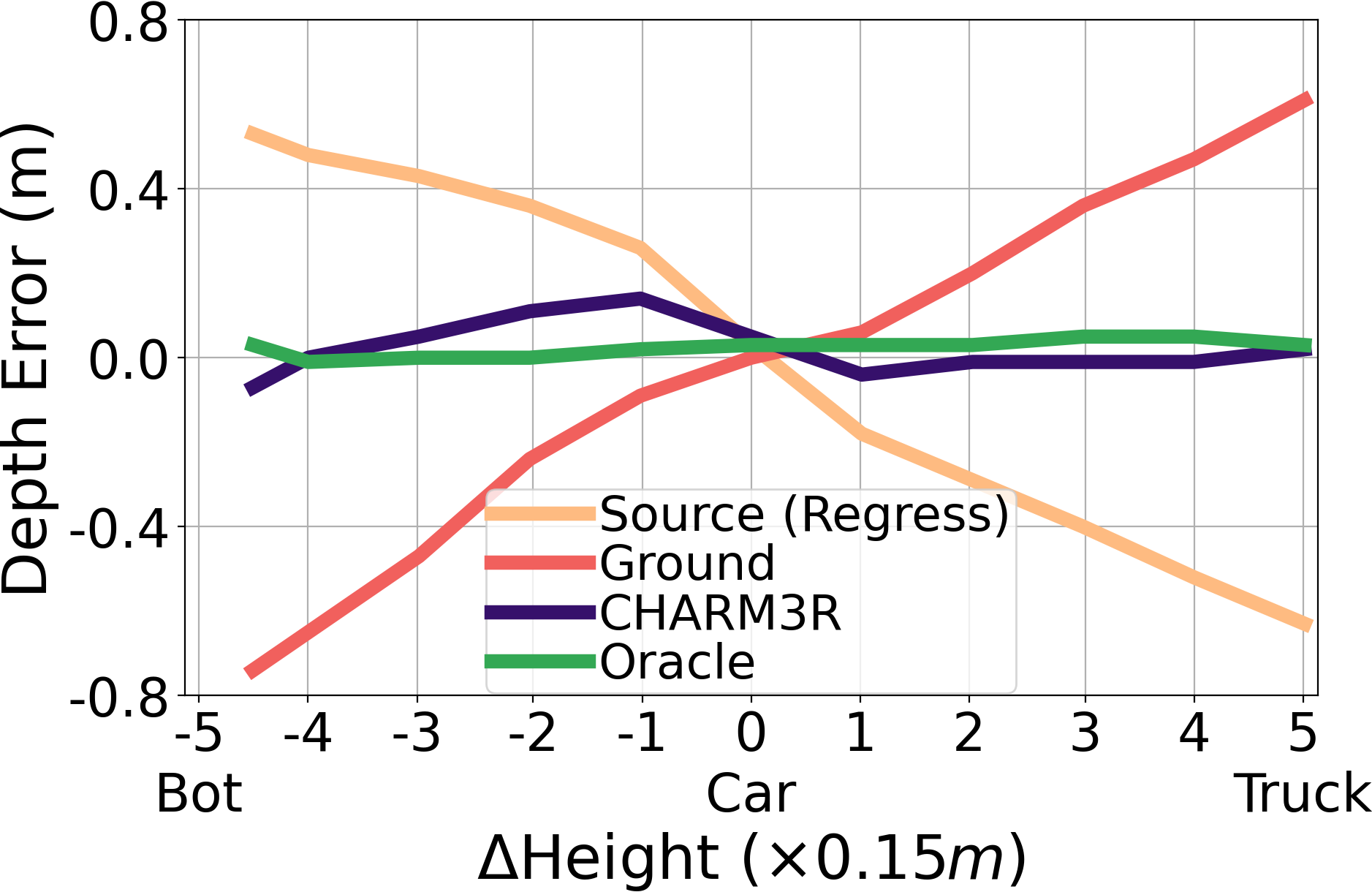}
                \caption{Depth error \trend on changing ego \variations.}
                \label{fig:reason_bias}
            \end{subfigure}
            \vspace{-1mm}
            \caption{\textbf{Performance Comparison.} The performance of \sota detector \gupNet \cite{lu2021geometry} drops significantly with changing ego \variations in inference.
            Ground-based model shows contrasting depth error (extrapolation) trend compared to regression-based depth models.
            Our proposed \textbf{\methodName exhibits greater robustness} to such variations by averaging regression and ground-based depth estimates. All methods, except the Oracle, are trained on car-height data $\egoHeightChange=0m$ and tested on data from bot to truck \variations.
           \vspace{-3mm} }
            \label{fig:reason}

        \end{figure*}

    The ego \variation of autonomous vehicles (AVs) varies significantly across platforms and deployment scenarios.
    While almost all training data are collected from a specific ego \variation, such as that of a passenger car, AVs are now deployed with substantially different ego \variations such as small bots or trucks.
    Collecting, labeling datasets and retraining models for each possible \variation is not scalable \cite{tzofi2023towards}, computationally expensive and impractical.
    Therefore, our work aims to address the challenge of \textit{generalizing} \monoThreeD models to \textit{unseen} ego \variations (See \cref{fig:teaser}).

    Generalizing \monoThreeD to unseen ego \variations from \textbf{single ego \variation} data is challenging due to the following five reasons.
    First, neural models excel at In-Domain (\inDomain{}) generalization, but struggle with unseen Out-Of-Domain (\outDomain{}) generalization \cite{xu2021how,teney2023id}.
    Second, ego \variation changes induce projective transformations \cite{hartley2003multiple} that CNNs \cite{cohen2016group}, \deviant \cite{kumar2022deviant} or \vit \cite{dosovitskiy2021image} backbones do not effectively handle \cite{sarkar2024shadows}.
    Third, existing projective equivariant backbones \cite{macdonald2022enabling,mironenco2024lie} are limited to single-transform-per-image scenarios, while every pixel in a driving image undergoes a different depth-dependent transform.
    Fourth, the non-linear projective transformations \cite{burns1992non,hartley2003multiple} makes interpolation difficult.
    Finally, disentangled learning does not work here since such approaches need at least two \variation data, while the training data here is from \textbf{single \variation}.
    Note that single height to multi height generalization is more practical since multi-height data is unavailable in almost all real datasets.

    We first systematically analyze and quantify the impact of ego \variation on the performance of \monoThreeD models trained on a single ego \variation.
    Leveraging the extended \carla dataset \cite{tzofi2023towards}, we evaluate the performance of state-of-the-art (\sota) \monoThreeD models under multiple ego \variations.
    Our analysis reveals that \sota \monoThreeD models exhibit significant performance degradation when faced with large \variation changes in inference (\cref{fig:reason_seventy,fig:reason_fifty}).
    Additionally, we empirically observe a consistent negative trend in the regressed object depth under \variation changes (\cref{fig:reason_bias}).
    Furthermore, we decompose the performance impact into individual sub-tasks and identify depth estimation as the primary contributor to this degradation.

    Recent works address ego \variation changes by using \plucker embeddings \cite{bahmani2024vd3d}, transforming target-height images to the original height, assuming constant depth \cite{li2025unidrive}, or by retraining with augmented data~\cite{tzofi2023towards}.
    While these techniques do offer some effectiveness, image transformation fails (\cref{fig:det_results_carla_val}) under significant \variation changes due to real-world depth variations.
    The augmentation strategy requires complicated pipelines for data synthesis at target heights and also falls short when the target \variation is \outDomain or when the target \variation is unknown apriori during training.

    To effectively generalize \monoThreeD to unseen ego \variations, a detector should first disentangle the depth representation from ego parameters in training and produce a new representation with new ego parameters in inference, while also canceling the \trends.
    We propose using the projected bottom \threeD center and ground depth in addition to the regressed depth.
    While the ground depth is easily calculated from ego parameters and height, and can be changed based on the ego \variation, its direct application to \monoThreeD models is sub-optimal (a reason why ground plane is not used alone).
    However, we observe a consistent positive trend in ground depth, which contrasts with the negative trend in regressed depths.
    By averaging both depth estimates within the model, we effectively cancel these opposing \trends and improve \monoThreeD generalization to unseen ego \variations.

    In summary the main contributions of this work include:
    \begin{itemize}
        \item We attempt the understudied problem of \outDomain ego \variation robustness in \monoThreeD models from single \variation data.
        \item We mathematically prove systematic negative and positive \trends in the regressed and ground-based object depths, respectively, with ego \variation changes under simplified assumptions (\cref{theorem:1,theorem:2}).
        \item We propose simple averaging of these depth estimates within the model to effectively counteract these opposing \trends and generalize to unseen ego \variations (\cref{sec:depth_merge}).
        \item We empirically demonstrate \sota robustness to unseen ego \variation changes on the \carla dataset (\cref{tab:det_results_carla_val}).
    \end{itemize}

\section{Literature Survey}\label{sec:literature}

    \noIndentHeading{Extrapolation / \outDomain Generalization.}
        Neural models excel at \inDomain generalization, but struggle at \outDomain generalization \cite{xu2021how,teney2023id}.
        There are two major classes of methods for good \outDomain classification.
        The first does not use target data and relies on diversifying data \cite{teney2021unshuffling}, features \cite{yashima2022feature,tiwari2023overcoming}, predictions \cite{lee2022diversify}, gradients \cite{ross2018learning,teney2022evading} or losses \cite{sagawa2019distributionally,ruan2023towards,puli2023don}.
        Another class finetunes on small target data \cite{kirichenko2022last}.
        None of these papers attempt \outDomain generalization for regression tasks.

    \noIndentHeading{Mono3D.}
        \monoThreeD has gained significant popularity, offering a cost-effective and efficient solution for perceiving the \threeD world.
        Unlike its more expensive \lidar and \radar counterparts \cite{riccardo-radar-hit-prediction-and-convolution-for-camera-radar-3d-object-detection,shi2019pointrcnn,yin2021center,long2023radiant}, or its computationally intensive stereo-based cousins \cite{Chen2020DSGN}, \monoThreeD relies solely on a single camera or multiple cameras with little overlaps.
        Earlier approaches to this task \cite{payet2011contours, chen2016monocular} relied on hand-crafted features, while the recent advancements use deep models.
        Researchers explored a variety of approaches to improve performance, including architectural innovations \cite{huang2022monodtr,xu2023mononerd}, equivariance \cite{kumar2022deviant, chen2023viewpoint}, losses \cite{brazil2019m3d,chen2020monopair}, uncertainty \cite{lu2021geometry,kumar2020luvli} and depth estimation \cite{zhang2021objects,min2023neurocs,yan2024monocd}.
        A few use NMS \cite{kumar2021groomed,liu2023monocular}, corrected extrinsics \cite{zhou2021monoef}, CAD models \cite{chabot2017deep, liu2021autoshape, lee2023baam} or \lidar \cite{reading2021categorical} in training.
        Other innovations include \pseudoLidar \cite{wang2019pseudo, ma2019accurate}, diffusion \cite{ranasinghe2024monodiff,xu20243difftection}, \bev feature encoding \cite{jiang2024fsd,zhang2022beverse,li2024bevnext} or transformer-based \cite{carion2020detr} methods with modified positional encoding \cite{shu2023dppe,tang2024simpb,hou2024open},  queries~\cite{li2023fast,zhang2023dabev,ji2024enhancing,chen2024learning} or query denoising \cite{liu2024ray}.
        Some use pixel-wise depth \cite{huang2021bevdet} or object-wise depth \cite{chu2023oabev,choi2023depth,liu2021voxel}.
        Many utilize temporal fusion with short \cite{wang2022sts,wang2023stream,liu2023petrv2,brazil2020kinematic} or long frame history \cite{park2022time,zong2023hop,changrecurrentbev} to boost performance.
        A few use distillation \cite{wang2023distillbev,kim2024labeldistill}, stereo \cite{wang2022sts,li2023bevstereo} or loss \cite{kumar2024seabird,liu2024multi} to improve these results further.
        For a comprehensive overview, we redirect readers to the surveys \cite{ma20233d,ma2022vision}.
        \methodName selects representative \monoThreeD models and improves their extrapolation to unseen camera \variations.

    \noIndentHeading{Camera Parameter Robustness.}
        While several works aim for robust \lidar-based detectors \cite{hu2022investigating,wang2020train,yang2021st3d,xu2021spg,chang2024cmda},  planners \cite{yao2024improving} and map generators \cite{ranganatha2024semvecnet}, fewer studies focus on generalizing image-based detectors.
        Existing image-based techniques, such as self-training \cite{li2022unsupervised}, adversarial learning \cite{wang2023towards}, perspective debiasing \cite{lu2023towards}, and multi-view depth constraints \cite{chang2024unified}, primarily address datasets with variations in camera intrinsics and minor height differences of $0.2m$.
        Some works show robustness to other camera parameters such as intrinsics \cite{brazil2023omni3d}, and rotations \cite{zhou2021monoef,moon2023rotation}.
        \methodName specifically tackles the challenge of generalizing to significant camera height changes, exceeding $0.7m$.

    \noIndentHeading{\Variation-Robustness.}
        Image-based \threeD detectors such as \bevHeight \cite{jia2023monouni} and MonoUNI \cite{jia2023monouni} train multiple detectors at different heights, but always do \inDomain testing.
        Recent works address ego \variation changes by either using \plucker embeddings \cite{bahmani2024vd3d,zhang2024camerasrays} for video generation/pose estimation, by transforming target-height images to the original height, assuming constant depth \cite{li2025unidrive} for \monoThreeD, or by retraining with augmented data~\cite{tzofi2023towards} for \bev segmentation.
        In contrast, we investigate the contrasting extrapolation behavior of regressed and ground-based depth estimators and average them for generalizing  \monoThreeD to unseen camera \variations.

    \noIndentHeading{Wide Baseline Setup.}
        Wide baseline setups are challenging due to issues like large occlusions, depth discontinuities \cite{strecha2003dense} and intensity variations \cite{strecha2004wide}.
        Unlike traditional wide-baseline setups with arbitrary baseline movements, generalization to unseen ego \variation requires handling baseline movements specifically along the vertical direction. 

\section{Notations and Preliminaries}

    We first list out the necessary notations and preliminaries which are used throughout this paper.
    These are not our contributions and can be found in the literature \cite{hartley2003multiple,garnett20193d,guo2020gen}.

\begin{figure}[!t]
        \centering
        \includegraphics[width=0.98\linewidth]{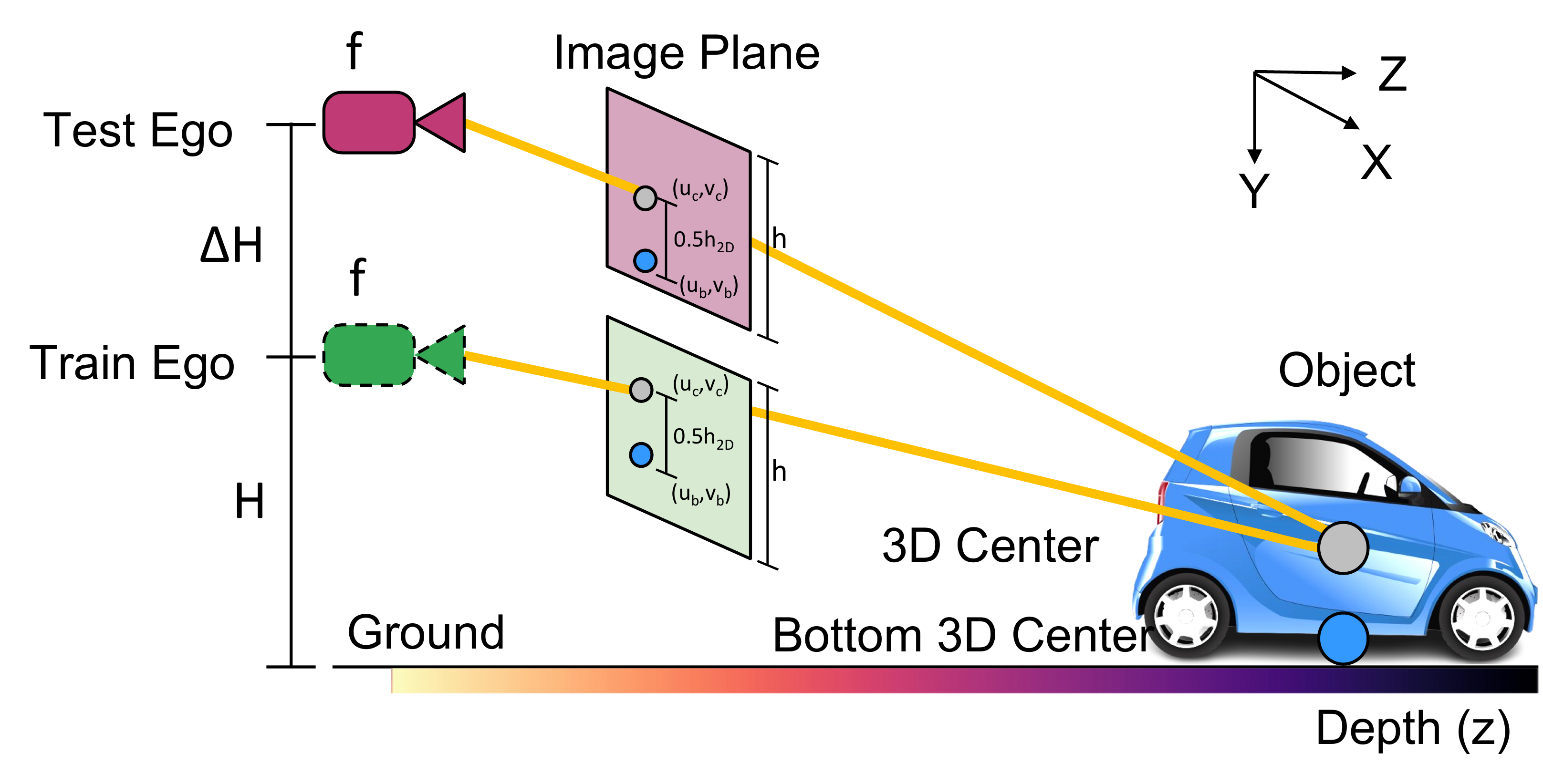}
        \vspace{-3mm}
        \caption{\textbf{Problem Setup}. Note that changing ego height does not change the object depth $\depthGT$ but only its position $(\pixUCenter, \pixVCenter)$ in the image plane. A regressed-depth model uses this pixel position to estimate the depth and therefore, fails when the ego height is changed.
        \vspace{-6mm}}
        \label{fig:setup}

    \end{figure}

    \noIndentHeading{Notations.}
        Let $\intrinsic\!\in\!\realDomain^{3 \times 3}$ denote the camera intrinsic matrix, $\rotation\!\in\!\realDomain^{3 \times 3}$ the rotation matrix and $\extrinsicTrans\!\in\!\realDomain^{3 \times 1}$ the translation vector of the extrinsic parameters.
        Also, $\zeroThreeD\!\in\!\realDomain^{3 \times 1}$ denotes the zero vector in \threeD.
        We denote the ego camera height on the car as $\egoHeight$, and the \variation change relative to this car as $\egoHeightChange$ meters.
        The camera intrinsics matrix $\intrinsic$ has focal length $\focal$ and principal point $(\ppointU, \ppointV)$.
        Let $(\pixU, \pixV)$ represent a pixel position in the camera coordinates, and
        $(\pixUCenter, \pixVCenter)$ and $(\pixUBottom, \pixVBottom)$ denotes the projected \threeD center and bottom center respectively.
        $\heightImage$ denotes the height of the image plane.
        We show these notations pictorially in \cref{fig:setup}.

    \begin{figure*}[!t]
        \centering
        \includegraphics[width=0.8\linewidth]{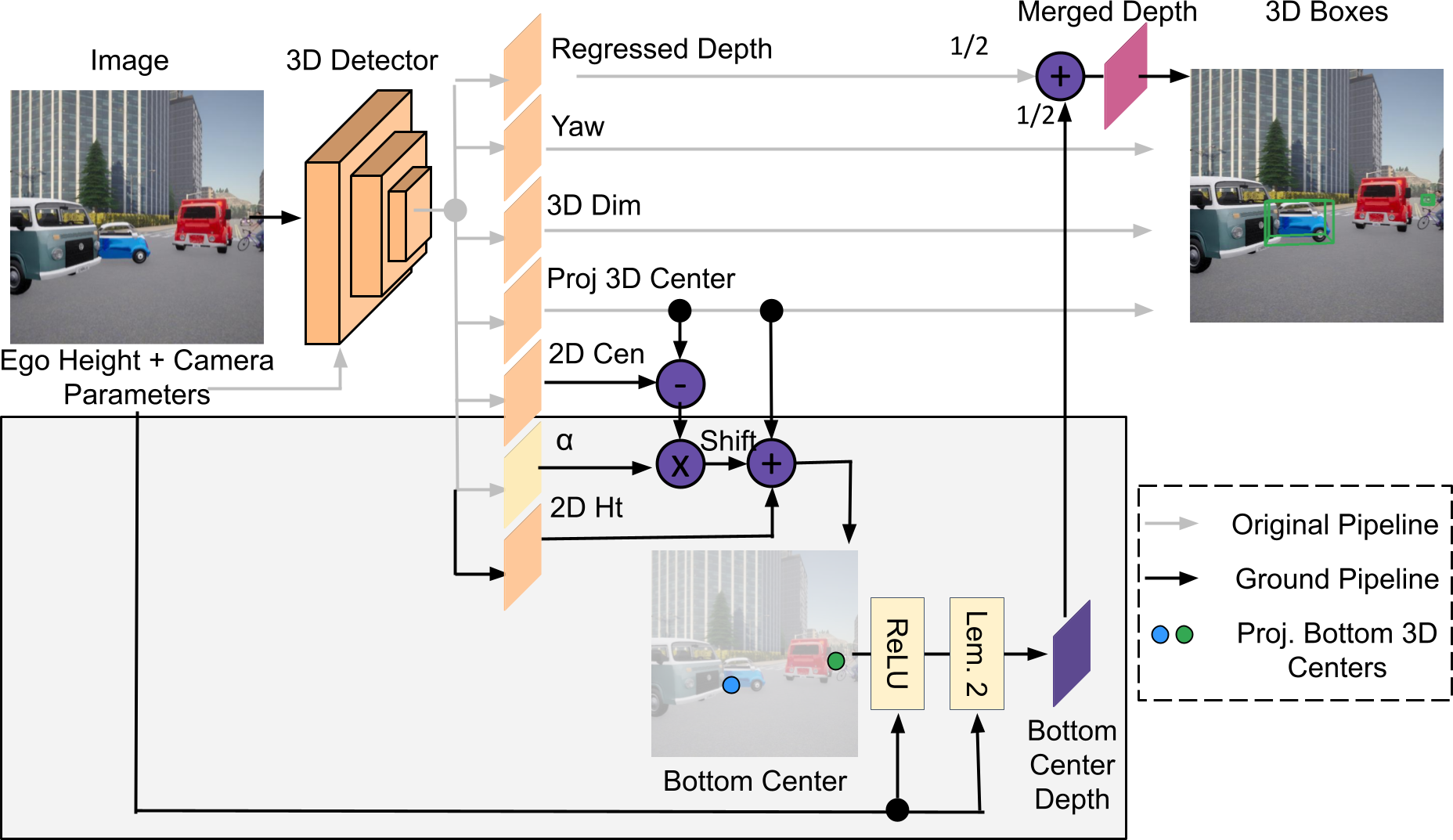}
        \caption{\textbf{Overview.} \methodName predicts the shift coefficient $\alpha$ to obtain projected \threeD bottom centers to query the ground depth and then averages the ground-depth and the regressed depth estimates within the model itself to output final depth estimate of a bounding box.
        \methodName uses \cref{theorem:1,theorem:2} to demonstrate that the ground and the regressed depth models show contrasting extrapolation behaviors.
        }
        \label{fig:overview}
        \vspace{-0.3cm}
    \end{figure*}

    \noIndentHeading{Pinhole Point Projection }\cite{hartley2003multiple}.
        The pinhole model relates a \threeD point $(\varX,\varY,\varZ)$ in the world coordinate system to its \twoD projected pixel $(\pixU, \pixV)$ in camera coordinates as:
        \begin{align}
        \begin{bmatrix}
        \pixU \\
        \pixV \\
        1 \\
        \end{bmatrix}\posZ
        =
        \begin{bmatrix}
        \intrinsic\;\;\;\zeroThreeD
        \end{bmatrix}
        \begin{bmatrix}
        \rotation\;\;\;\extrinsicTrans \\
        \zeroThreeD^T\;\;1
        \end{bmatrix}
        \begin{bmatrix}
        \varX \\
        \varY \\
        \varZ \\
        1
        \end{bmatrix},
        \label{eq.pinhole}
        \end{align}
        where $\posZ$ denotes the depth of pixel $(\pixU, \pixV)$.

    \noIndentHeading{Ground Depth Estimation }\cite{garnett20193d,guo2020gen}.
        While depth estimation in \monoThreeD is ill-posed, ground depth is precisely determined given the camera parameters and height relative to the ground in the world coordinate system \cite{garnett20193d,guo2020gen,yang2023gedepth}.
        Since all datasets provide camera mounting height, we obtain the depth of ground plane pixels in the closed form.

        \begin{lemma}\label{lemma:1}
            \textbf{Ground Depth of Pixel }\cite{garnett20193d,guo2020gen,yang2023gedepth}.
            Consider a pinhole camera model with intrinsics $\intrinsic$, rotation $\rotation$ and translation extrinsics $\extrinsicTrans$. Let matrix $\bm{A}\!=\!(a_{ij})\!=\!\rotation^{-1}\intrinsic^{-1}\in \realDomain^{3 \times 3}$, and $-\rotation^{-1}\extrinsicTrans$ as the vector $\bm{B}\!=\!(b_i) \in \realDomain^{3 \times 1}$. Then, the ground depth $\posZ$ for a pixel $(\pixU, \pixV)$ is
            \begin{align}
                \posZ = \frac{\egoHeight - b_{2}}{a_{21}\pixU + a_{22}\pixV + a_{23}}.
                \label{eq:gd_depth}
            \end{align}
        \end{lemma}
        We refer to \cref{sec:proof_lemma} in the appendix for the derivation and \cref{sec:not_parallel,sec:not_flat} for extensions to non-parallel and non-flat roads respectively.

        \begin{lemma}\label{lemma:2}
        \textbf{Ground Depth of Pixel }For datasets with the rotation extrinsics $\rotation$ an identity, the depth estimate $\posZ$ from \cref{lemma:1} becomes
        \begin{align}
            \posZ &= \dfrac{\egoHeight - b_{2}}{~\dfrac{\pixV\!-\!\ppointV}{\focal}~}.
            \label{eq:gd_depth_simple}
        \end{align}
        \end{lemma}
        We refer to \cref{sec:proof_simple} for the proof.

\section{\methodName}\label{sec:method}

    In this section, we first mathematically prove the contrasting extrapolation behavior of regressed and ground-based object depths under varying camera \variations.
    To mitigate the impact of these opposing \trends and improve generalization to unseen \variations, we propose \methodNameFull or \methodName.
    \methodName averages both these depth estimates within the model to mitigate these \trends and improves generalization to unseen \variations.
    \cref{fig:overview} shows the overview of \methodName.

    \subsection{Ground-based Depth Model}\label{sec:bottom_ground_depth}
        Outdoor driving scenes typically contain a ground region, unlike indoor scenes.
        The ground depth varies with ego \variation, providing a valuable reference and prior for generalizing \monoThreeD to unseen ego \variations.

        \noIndentHeading{Bottom Center Estimation.}
            \cref{lemma:1} utilizes the ground plane depth from \cref{eq:gd_depth} to estimate object depths.
            The numerator in \cref{eq:gd_depth} can be negative, while depth is positive for forward facing cameras.
            To ensure positive depth values, we apply the Rectified Linear Unit (\relu) activation $(\max(\posZ, 0))$ to the numerator of \cref{eq:gd_depth}.
            This step promotes spatially continuous and meaningful ground depth representations, improving the training stability of \methodName.
            Ablation in \cref{sec:ablation} confirm the effectiveness.

            In practice, \methodName leverages the projected \threeD center $(\pixUCenter,\pixVCenter)$, \twoD height information $\heightBoxTwoD$ and the \twoD center  $(\pixUCenterTwoD,\pixVCenterTwoD)$ to compute the projected bottom \threeD center $(\pixUBottom,\pixVBottom)$ as follows:
            \begin{align}
                \pixUBottom &= \pixUCenter~~; \quad
                \pixVBottom = \pixVCenter + \frac{1}{2}\heightBoxTwoD + \alpha (\pixVCenter - \pixVCenterTwoD).
                \label{eq:bottom_center}
            \end{align}

            With the projected bottom center $(\pixUBottom,\pixVBottom)$ estimated, we query the ground plane depth at this point, as derived in \cref{lemma:2}.
            Note that we do not use the \threeD height to calculate the bottom center since projecting this point requires the box depth, which is the quantity we aim to estimate.
            The learnable correction factor $\alpha$ compensates for the perspective effects to map the bottom center to the projected box center.
            We now analyze the extrapolation behavior of this ground-based depth model in the following theorem.

        \begin{theorem}\label{theorem:1}
            \textbf{Ground-based bottom center model has positive slope (trend) in extrapolation.}
            Consider a ground depth model that predicts $\depthPred$ from the projected bottom \threeD center $(\pixUBottom,\pixVBottom)$ image.
            Assuming the GT object depth $\depthGT$ is more than the ego height change $\egoHeightChange$, the mean depth error of the ground model exhibits a positive \trend \wrt~the \variation change $\egoHeightChange$:
            \begin{align}
                \label{eqn:ground_depth_bias}
                \expect\Big(\newDepthGround - \depthGT\Big) &\approx
                \relu\left(\frac{1}{\pixVBottom\!-\!\ppointV}\right)\focal\egoHeightChange,
            \end{align}
            where $\focal$ is the focal length and $(\ppointU,\ppointV)$ is the optical center.
        \end{theorem}
        \cref{theorem:1} says that the ground model over-estimates and under-estimates depth as the ego \variation change $\egoHeightChange$ increases and decreases respectively.
        \begin{proof}
            With camera shift of $\egoHeightChange{}m$, the $y$-coordinate of the projected \threeD bottom center $\pixVBottom$ of a \threeD box becomes $\pixVBottom\!+\! \dfrac{\focal\egoHeightChange}{\depthGT}$ (\cref{sec:pix_shift}).
            Using \cref{eq:gd_depth_simple}, the new depth $\newDepthGround$ is
            \begin{align}
                \newDepthGround = \dfrac{\egoHeight + \egoHeightChange - b_{2}}{\dfrac{~\pixVBottom + \dfrac{\focal\egoHeightChange}{\depthGT} - \ppointV}{\focal}~}
                &= \dfrac{\egoHeight + \egoHeightChange - b_{2}}{~\dfrac{\pixVBottom\!-\!\ppointV}{\focal} + \dfrac{\egoHeightChange}{\depthGT}~}.
            \end{align}
            If the ego height change $\egoHeightChange$ is small compared to the object depth $\depthGT$, $\dfrac{\egoHeightChange}{\depthGT} \approx 0$.
            So, we write the above equation as
            \begin{align}
                \newDepthGround &\approx \dfrac{\egoHeight + \egoHeightChange - b_{2}}{\dfrac{\pixVBottom\!-\!\ppointV}{\focal}} = \oldDepthGround + \dfrac{\egoHeightChange}{~\dfrac{\pixVBottom\!-\!\ppointV}{\focal}~} \nonumber \\
                &\approx \depthGT + \noise + \dfrac{\focal\egoHeightChange}{\pixVBottom\!-\!\ppointV} \nonumber \\
                \implies \newDepthGround - \depthGT &\approx \noise + \dfrac{\focal\egoHeightChange}{\pixVBottom\!-\!\ppointV} \nonumber,
            \end{align}
            assuming the ground depth $\oldDepthGround$ at train height $\egoHeightChange=0$ is the GT depth $\depthGT$ added by a normal random variable $\noise$ with mean $0$ and variance $\normalVar$ as in \cite{kumar2024seabird}.
            Taking expectation on both sides, the mean depth error is
            \begin{align}
                \expect\Big(\newDepthGround - \depthGT\Big) &\approx
                \left(\dfrac{1}{\pixVBottom\!-\!\ppointV}\right)\focal\egoHeightChange, \nonumber
            \end{align}
            confirming the positive \trend of the mean depth error of the ground model \wrt~the height change $\egoHeightChange$.

            The ground lies between the bottom part of the image plane/ image height ($h$) and the optical center $y$-coordinate $\ppointV$, and so $\pixVBottom-\ppointV > 0$.
            However, in practice, it could get negative in early stage of training.
            To enforce non-negativity of this term, we pass $\pixVBottom\!-\!\ppointV$ through a \relu non-linearity to enforce $\pixVBottom\!-\!\ppointV$ is positive.
            \cref{sec:ablation} confirms that \relu remains important for good results.
        \end{proof}

    \subsection{Regression-based Depth Model}
        Most \monoThreeD models rely on regression losses, to compare the predicted depth with the GT depth~\cite{kumar2022deviant, zhang2022beverse}.
        We, next, derive the extrapolation behavior of such regressed depth model in the following theorem.

        \begin{theorem}\label{theorem:2}
            \textbf{Regressed model has negative slope (trend) in extrapolation}.
            Consider a regressed depth model trained on data from a single ego \variation, predicting depth $\depthPred$ from the projected \threeD center $(\pixUCenter,\pixVCenter)$.
            Assuming a linear relationship between predicted depth and pixel position, the mean depth error of a regressed model exhibits a negative \trend \wrt~the \variation change $\egoHeightChange$:
            \begin{align}
                \expect\Big(\newDepthRegress - \depthGT\Big) &= -\left(\dfrac{\depthSlope}{\depthGT}\right)\focal\egoHeightChange,
                \label{eqn:regress_depth_bias}
            \end{align}
            where $\depthSlope$ is a camera \variation independent positive constant.
        \end{theorem}

        \begin{figure}[!t]
            \centering
            \includegraphics[width=\linewidth]{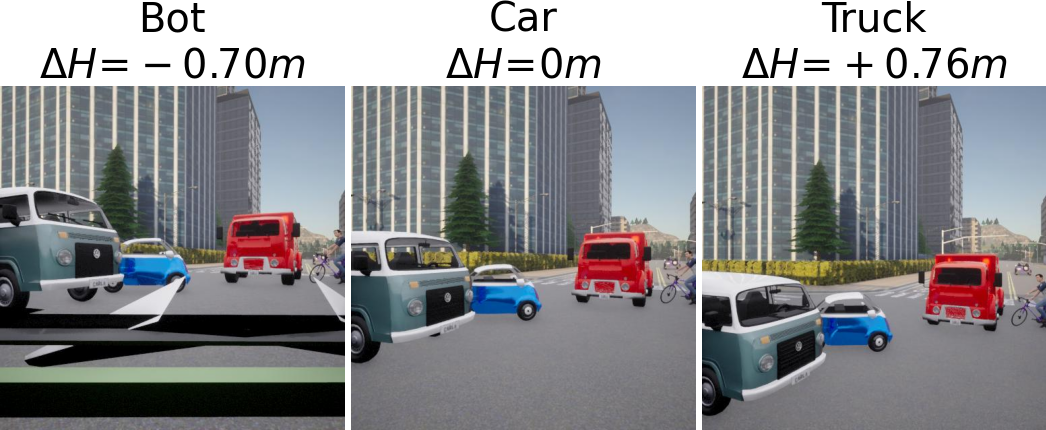}
            \includegraphics[width=\linewidth]{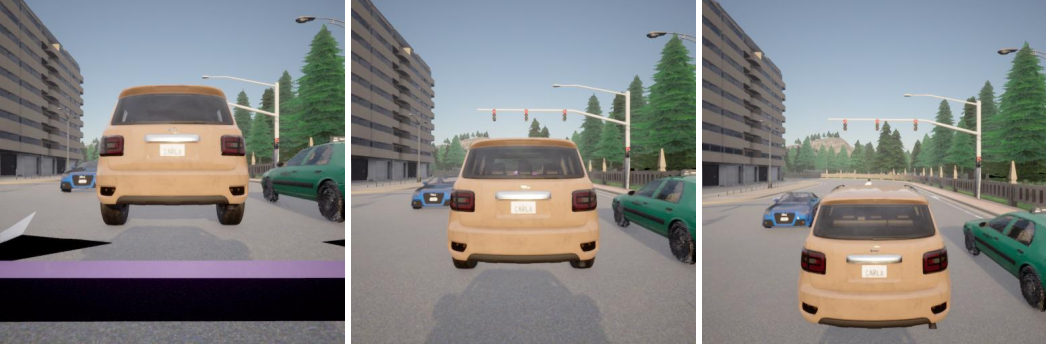}
            \caption{\textbf{\carla \val samples} with both negative and positive ego \variation changes ($\egoHeightChange$) covers AVs from bots to cars to trucks.}
            \label{fig:carla_sample}
            \vspace{-0.5cm}
        \end{figure}

        \cref{theorem:2} says that regressed depth model under-estimates and over-estimates depth as the ego \variation change $\egoHeightChange$ increases and decreases respectively.

        \begin{table*}[!t]
            \centering
            \scalebox{\scaleFraction}{
            \setlength\tabcolsep{0.23cm}
            \begin{tabular}{ccccccc m dcd m dcd m dcdc }
                \addlinespace[0.01cm]
                \multicolumn{7}{cm}{Oracle Params. \downarrowRHDSmall~~/ $\egoHeightChange~(m)\rightarrowRHDSmall$ } &  \multicolumn{3}{cm}{\apThreeDSeventy~\bracketPercentage~(\uparrowRHDSmall)} & \multicolumn{3}{cm}{\apThreeDFifty~\bracketPercentage~(\uparrowRHDSmall)} & \multicolumn{3}{c}{\MDE $(m)~[\approx 0]$}\\
                $x$ & $y$ & $z$ & $l$ & $w$ & $h$ & $\theta$ & $-0.70$ & $0$ & $+0.76$ & $-0.70$ & $0$ & $+0.76$ & $-0.70$ & $0$ & $+0.76$ \\
                \myTopRule
                & & & & & & & $9.46$ & $53.82$ & $7.23$ & $41.66$ & $76.47$ & $40.97$ & $+0.53$ & $+0.03$ & $-0.63$\\
                \cmark & & & & & & & $15.95$ & $62.21$ & $12.74$ & $46.89$ & $76.78$ & $50.97$ & $+0.53$ & $+0.03$ & $-0.63$\\
                & \cmark  && & & & & $13.56$ & $59.55$ & $10.67$ & $44.93$ & $76.86$ & $49.84$ & $+0.53$ & $+0.03$ & $-0.63$\\
                &&  \cmark & & & & & $34.82$ & $69.99$ & $39.03$ & $68.10$ & $82.73$ & $76.24$ & $+0.00$ & $+0.00$ & $+0.00$\\
                \cmark & \cmark & \cmark & & & & & $65.44$ & $82.36$ & $80.70$ & $74.76$ & $84.93$ & $82.11$ & $+0.00$ & $+0.00$ & $+0.00$\\
                & & & \cmark & \cmark & \cmark & & $10.32$ & $56.24$ & $7.20$ & $42.04$ & $76.61$ & $42.03$ & $+0.53$ & $+0.03$ & $-0.63$\\
                \cmark & \cmark & \cmark & \cmark & \cmark & \cmark & & $75.86$ & $82.82$ & $82.08$ & $78.21$ & $85.17$ & $82.24$ & $+0.00$ & $+0.00$ & $+0.00$\\
                \cmark & \cmark & \cmark & \cmark & \cmark & \cmark & \cmark & $78.44$ & $85.20$ & $82.28$ & $78.44$ & $85.20$ & $82.28$ & $+0.00$ & $+0.00$ & $+0.00$\\
            \end{tabular}
            }
            \caption{\textbf{Error analysis} of \gupNet \cite{lu2021geometry} trained on $\egoHeightChange\!=\!0m$ on all height changes $\egoHeightChange$ of \carla \val split.
            Depth remains the biggest source of error in inference on \hl{unseen ego \variations}.
            }
            \label{tab:error_analysis}
            \vspace{-0.5cm}
        \end{table*}

        \begin{proof}
            Neural nets often use the $y$-coordinate of their projected \threeD center $\pixVCenter$ to predict depth \cite{dijk2019neural}.
            Consider a simple linear regression model for predicting depth.
            Then, the regressed depth $\oldDepthRegress$ is
            \begin{align}
                \oldDepthRegress &= -\left(\dfrac{\depthGTMax\!-\!\depthGTMin}{\heightImage\!-\!\ppointV}\right)(\pixVCenter\!-\!\ppointV) + \depthGTMax \nonumber \\
                \implies \oldDepthRegress &= -\depthSlope(\pixVCenter\!-\!\ppointV) + \depthGTMax.
                \label{eq:regress_depth}
            \end{align}

            This linear regression model has a negative slope, with a positive slope parameter $\depthSlope$, and $\heightImage$ being the height of the image.
            This model predicts depth $\depthGTMin$ at pixel position $\pixVCenter\!=\!\heightImage$ and $\depthGTMax$ at principal point $\pixVCenter\!=\!\ppointV$.
            With ego shift of $\egoHeightChange{}m$, the projected center of the object becomes $\pixVCenter + \dfrac{\focal\egoHeightChange}{\depthGT}$ (\cref{sec:pix_shift}).
            Substituting this into the regression model of \cref{eq:regress_depth}, we obtain the new depth $\newDepthRegress$ as,
            \begin{align}
                \newDepthRegress &= -\depthSlope\left(\pixVCenter\!+\! \dfrac{\focal\egoHeightChange}{\depthGT}\!-\!\ppointV\right) + \depthGTMax \nonumber \\
                &= -\depthSlope(\pixVCenter\!-\!\ppointV) + \depthGTMax -\left(\dfrac{\depthSlope}{\depthGT}\right)\focal\egoHeightChange \nonumber \\
                &= \oldDepthRegress -\left(\dfrac{\depthSlope}{\depthGT}\right)\focal\egoHeightChange \nonumber \\
                &= \depthGT +  \noise -\left(\dfrac{\depthSlope}{\depthGT}\right)\focal\egoHeightChange \nonumber \\
                \implies \newDepthRegress - \depthGT &= \noise -\left(\dfrac{\depthSlope}{\depthGT}\right)\focal\egoHeightChange \nonumber,
            \end{align}
            assuming the regressed depth $\oldDepthRegress$ at train height $\egoHeightChange=0$ is the GT depth $\depthGT$ added by a normal random variable $\noise$ with mean $0$ and variance $\normalVar$ as in \cite{kumar2024seabird}.
            Taking expectation on both sides, the mean depth error is
            \begin{align}
                \expect\Big(\newDepthRegress - \depthGT\Big) &= -\left(\dfrac{\depthSlope}{\depthGT}\right)\focal\egoHeightChange, \nonumber
            \end{align}
            confirming the negative \trend of the mean depth error of the regressed depth model \wrt~the height change $\egoHeightChange$.
        \end{proof}

    \subsection{Merging Depth Estimates}\label{sec:depth_merge}

        \cref{theorem:1,theorem:2} prove that the ground and the regressed depth models show contrasting extrapolation behaviors.
        The former over-estimates the depth while the latter under-estimates depth as the ego \variation change $\egoHeightChange$ increases.
        \cref{fig:overview} shows how these two depth estimates are fused together.
        Overall, \methodName leverages depth information from these two source sources (with different extrapolation behaviors) to improve the \monoThreeD generalization to unseen camera \variations.
        \methodName starts with an input image, and estimates the depth of the object using two methods: ground and regressed depth.
        \methodName outputs the projected bottom center of the object to query the ground depth (calculated from the ego camera parameters and its position and orientation relative to the ground plane as in \cref{lemma:1}).
        It also outputs another depth estimate based on regression.
        The final step combines the two estimated depths with a simple average to cancel the opposing trends and obtain the refined depth estimates, resulting in a set of accurate and localized \threeD objects in the scene.

\section{Experiments}\label{sec:experiments}

    \noIndentHeading{Datasets.}
        Our experiments utilize the simulated \carla dataset\footnote{The authors of \cite{tzofi2023towards} do not release their other Nvidia-Sim dataset.} from \cite{tzofi2023towards}, configured to mimic the \nuscenes~\cite{caesar2020nuscenes} dataset.
            We use this dataset for two reasons.
            First, this dataset reduces training and testing domain gaps, while existing public datasets lack data at multiple ego \variations.
            Second, recent work \cite{tzofi2023towards} also uses this dataset for their experiments.
            The default \carla dataset sweeps camera height changes $\egoHeightChange$ from $0$ to $0.76m$, rendering a dataset every $0.076m$ (car to trucks).
            To fully investigate the impact of camera height variations, we extend the original \carla dataset by introducing negative height changes.
            The extended \carla dataset sweeps height changes $\egoHeightChange$ from $-0.70m$ to $0.76m$ with settings from bots to cars to trucks.
            \cref{fig:carla_sample} illustrates sample images from this dataset.
            Note that we exclude $\egoHeightChange\!=\!-0.76m$ setting due to visibility obstructions caused by the ego vehicle's bonnet.

        \begin{table*}[!t]
            \centering
            \scalebox{\scaleFraction}{
            \setlength\tabcolsep{0.23cm}
            \begin{tabular}{l l m dcd m dcd m dcdc}
                \addlinespace[0.01cm]
                \multirow{2}{*}{\threeD Detector} & \multirow{2}{*}{Method $\downarrowRHDSmall$ / $\egoHeightChange~(m)\rightarrowRHDSmall$} & \multicolumn{3}{cm}{\apThreeDSeventy \bracketPercentage~(\uparrowRHDSmall)} & \multicolumn{3}{cm}{\apThreeDFifty \bracketPercentage~(\uparrowRHDSmall)} & \multicolumn{3}{c}{\MDE $(m)~[\approx 0]$}\\
                &  & $-0.70$ & $0$ & $+0.76$ & $-0.70$ & $0$ & $+0.76$ & $-0.70$ & $0$ & $+0.76$\\
                \myTopRule
                \multirow{6}{*}{\gupNet \cite{lu2021geometry}}
                & Source  & $9.46$ & $53.82$ & $7.23$ & $41.66$ & $76.47$ & $40.97$ & $+0.53$ & $+0.03$ & $-0.63$\\
                & \plucker \cite{plucker1828analytisch} & $8.43$ & $55.56$ & $10.13$ & $37.10$ & \first{76.57} & $43.22$ & $+0.55$ & $+0.03$ & $-0.63$\\
                & \uniDrive \cite{li2025unidrive} & $10.73$ & $53.82$ & $5.54$ & $42.30$ & $76.46$ & $39.33$ & $+0.51$ & $+0.03$ & $-0.67$\\
                & \uniDrivePlus \cite{li2025unidrive} & $10.83$ & $53.82$ & $12.27$ & $47.81$ & $76.46$ & $53.08$ & $+0.39$ & $+0.03$ & $-0.48$\\
                & \cellcolor{methodColor}\textbf{\methodName} & \cellcolor{methodColor}\first{19.45} & \cellcolor{methodColor}\first{55.68} &  \cellcolor{methodColor}\first{27.33} & \cellcolor{methodColor}\first{53.40} & \cellcolor{methodColor}${74.47}$ & \cellcolor{methodColor}\first{61.98} & \cellcolor{methodColor}$+0.07$ & \cellcolor{methodColor}$+0.05$ & \cellcolor{methodColor}$-0.02$ \\
                \hhline{|~|-----------|}
                & Oracle & $70.96$ & $53.82$ & $62.25$ & $83.88$ & $76.47$ & $83.96$ & $+0.03$ & $+0.03$ & $+0.03$\\
                \myTopRule
                \multirow{6}{*}{\deviant \cite{kumar2022deviant}}
                & Source  & $8.63$ & $50.18$ & $6.25$ & $40.24$ & $73.78$ & $41.74$ & $+0.46$ & $+0.01$ & $-0.65$ \\
                & \plucker \cite{plucker1828analytisch} & $8.43$ & \first{51.32} & $9.52$ & $38.24$ & \first{73.91} & $44.22$ & $+0.46$ & $+0.01$ & $-0.64$ \\
                & \uniDrive \cite{li2025unidrive} & $8.33$ & $50.18$ & $6.56$ & $41.40$ & $73.78$ & $41.27$ & $+0.46$ & $+0.01$ & $-0.64$ \\
                & \uniDrivePlus \cite{li2025unidrive} & $6.73$ & $50.18$ & $12.03$ & $42.91$ & $73.78$ & $52.36$ & $+0.37$ & $+0.01$ & $-0.47$ \\
                & \cellcolor{methodColor}\textbf{\methodName} & \cellcolor{methodColor}\first{17.11} & \cellcolor{methodColor}$48.74$ & \cellcolor{methodColor}\first{26.24} & \cellcolor{methodColor}\first{49.28} & \cellcolor{methodColor}${70.21}$ & \cellcolor{methodColor}\first{63.60} & \cellcolor{methodColor}$+0.01$ & \cellcolor{methodColor}$+0.03$ & \cellcolor{methodColor}$-0.02$  \\
                \hhline{|~|-----------|}
                & Oracle & $71.97$ & $50.18$ & $62.56$ & $84.56$ & $73.78$ & $83.94$ & $+0.03$ & $+0.01$ & $-0.02$ \\
            \end{tabular}
            }
            \caption{\textbf{\carla \val Results.}
            \methodName \textbf{outperforms} all other baselines, especially at bigger \hl{unseen ego \variations}.
            All methods except Oracle are trained on car height $\egoHeightChange=0m$ and tested on bot to truck height data.
            [Key: \firstKey{Best}]
            }
            \label{tab:det_results_carla_val}
        \end{table*}
        \begin{figure*}[!t]
            \centering
            \begin{subfigure}{.3\linewidth}
                \includegraphics[width=\linewidth]{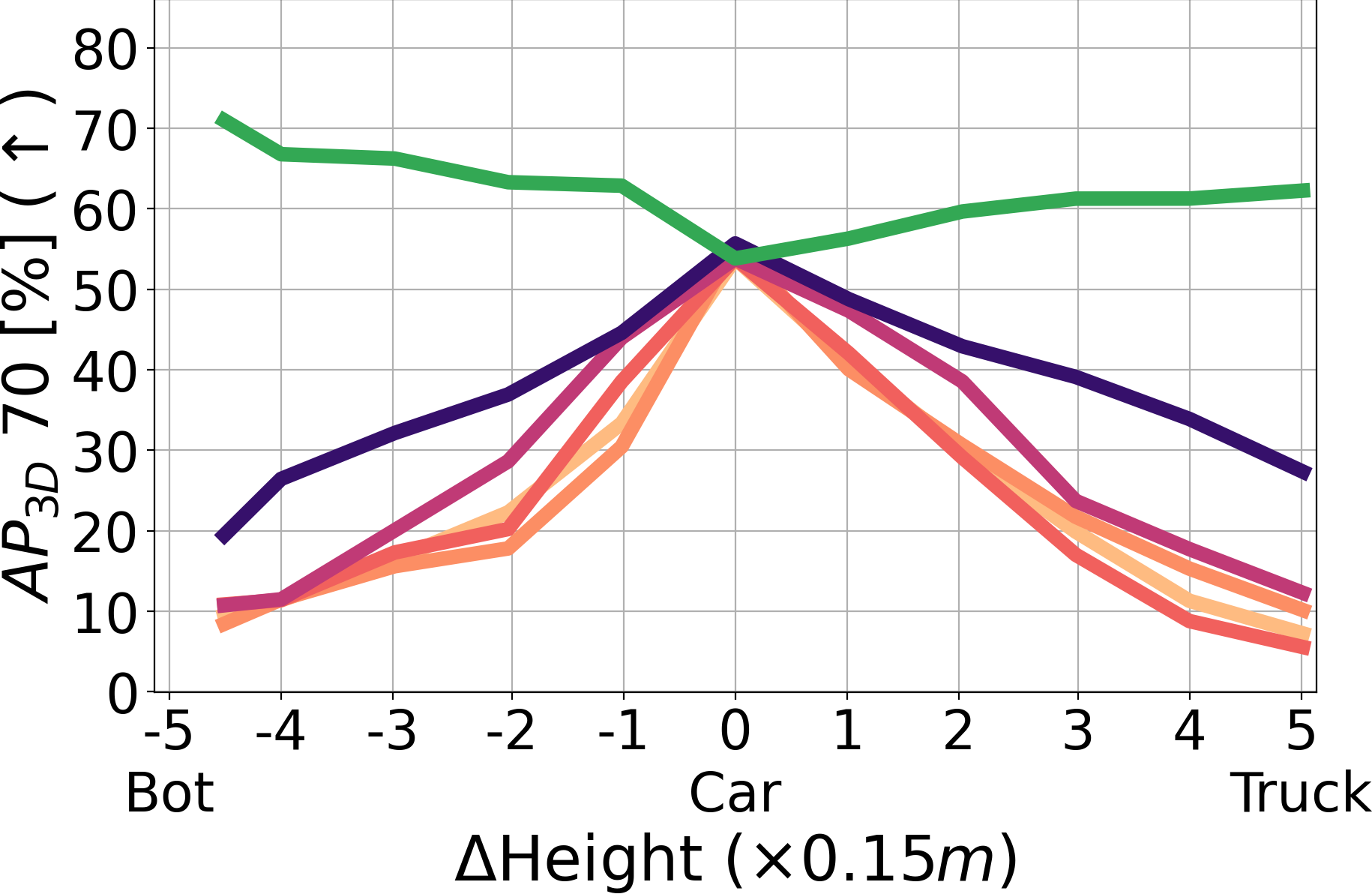}
                \caption{\apThreeDSeventy \bracketPercentage{} comparison.}
            \end{subfigure}%
            \hfill
            \begin{subfigure}{.3\linewidth}
                \includegraphics[width=\linewidth]{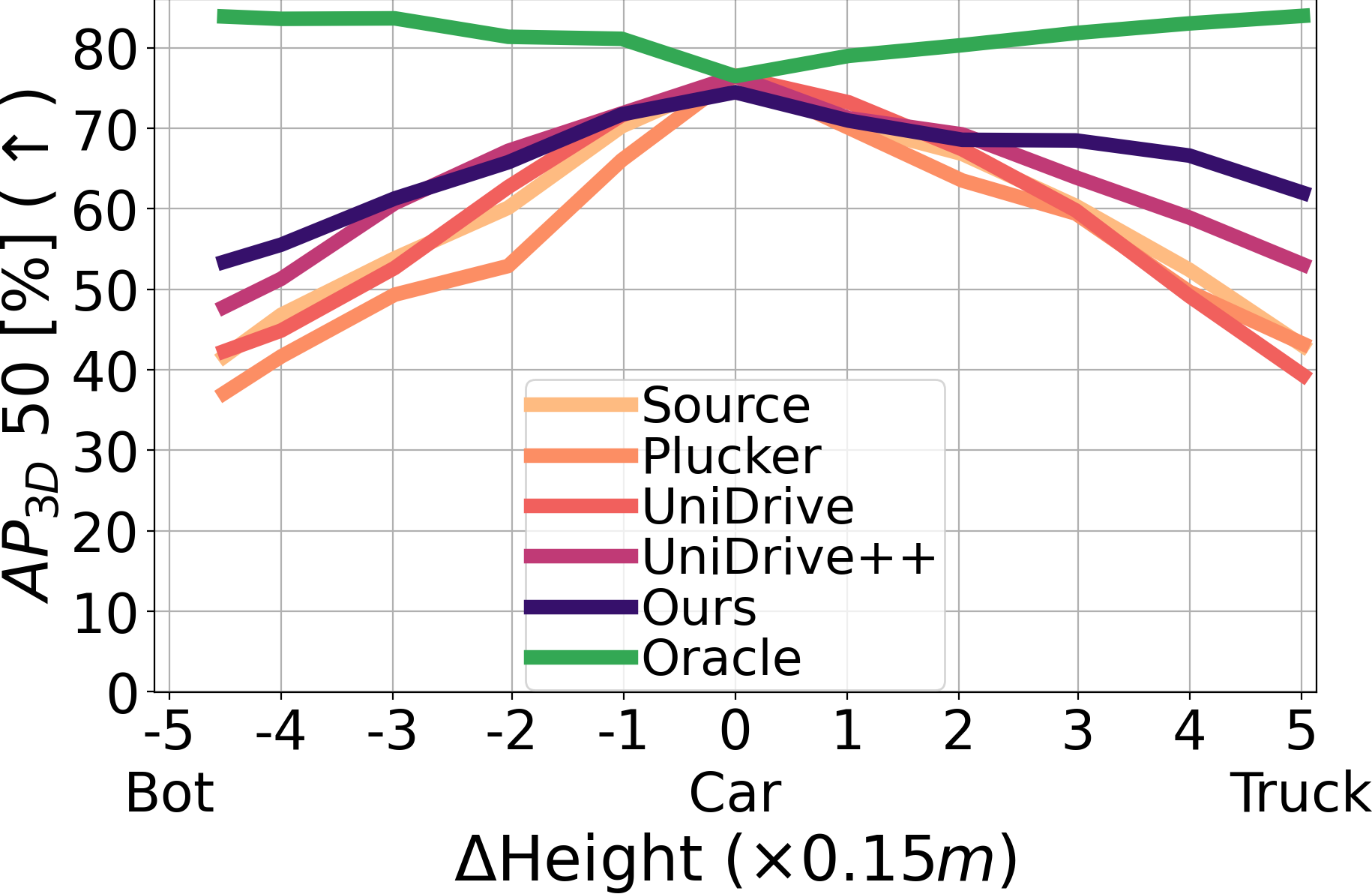}
                \caption{\apThreeDFifty \bracketPercentage{} comparison.}
            \end{subfigure}
            \hfill
            \begin{subfigure}{.3\linewidth}
                \includegraphics[width=\linewidth]{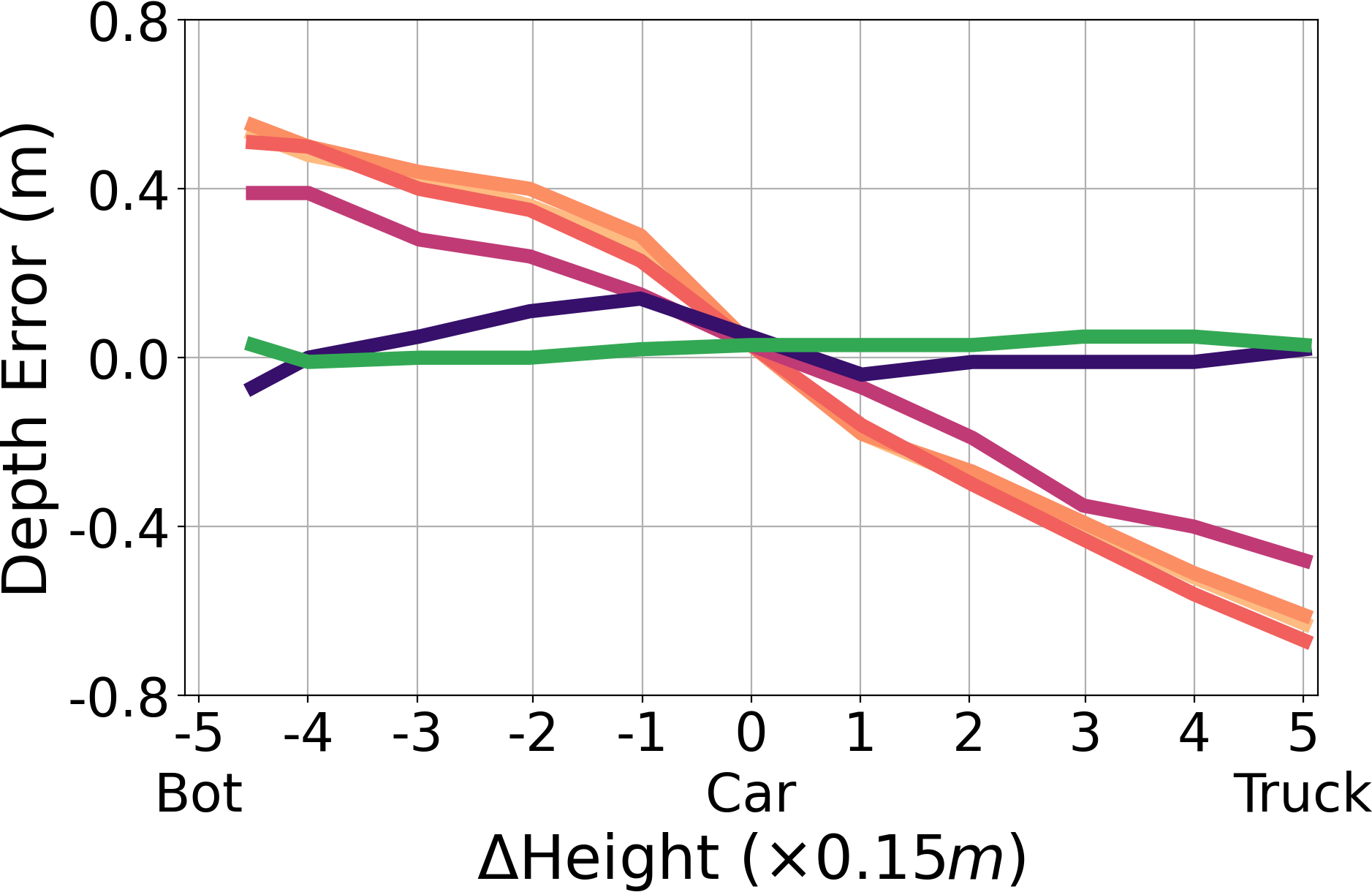}
                \caption{\MDE comparison.}
                \label{fig:mde_results_carla_val_gup}
            \end{subfigure}
            \caption{\textbf{\carla \val Results on \gupNet.}
            \methodName \textbf{outperforms} all baselines, especially at bigger {unseen ego \variations}.
            All methods except Oracle are trained on car height and tested on all \variations.
            Results of inference on height changes of $-0.70,0$ and $0.76$ meters are in \cref{tab:det_results_carla_val}.
            See \cref{fig:det_results_carla_val} in the supplementary for another detector.
            }
            \label{fig:det_results_carla_val}
        \end{figure*}

    \noIndentHeading{Data Splits.}
        Our experiments use the 
            \textit{\carla \val Split.}
            This dataset split \cite{tzofi2023towards} contains $25{,}000$ images ($2{,}500$ scenes) from \textit{town03} map for training and $5{,}000$ images ($500$ scenes) from \textit{town05} map for inference on multiple ego \variations. Except for Oracle, we train all models on training images from the car height $(\egoHeightChange=0m)$.

    \noIndentHeading{Evaluation Metrics.}
        We choose the \kitti \apThreeDSeventy percentage on the Moderate category \cite{geiger2012we} as our evaluation metric.
        We also report \apThreeDFifty percentage numbers following prior works \cite{brazil2020kinematic,kumar2021groomed}.
        Additionally, we report the mean depth error (\MDE) over predicted boxes with \iouTwoD overlap greater than $0.7$ with the GT boxes similar to \cite{kumar2022deviant}.
        Note that \MDE is different from MAE metric of \cite{kumar2022deviant} that it does not take absolute value.

    \noIndentHeading{Detectors.}
        We use the \gupNet \cite{lu2021geometry} and \deviant \cite{kumar2022deviant} as our base detectors.
        The choice of these models encompasses CNN \cite{lu2021geometry} and group CNN-based \cite{kumar2022deviant} 
        architectures.

    \noIndentHeading{Baselines.}
        We compare against the following:
        \begin{itemize}
            \item \textit{Source}: This is the \monoThreeD model trained on the car height ($\egoHeightChange=0m$) data.
            \item \textit{\plucker Embeddings} \cite{plucker1828analytisch,teller1999determining}: Training a \monoThreeD model with \plucker embeddings to improve robustness as in \threeD pose estimation and reconstruction tasks. \plucker embeddings generalize the intrinsic-focused \camConvs \cite{facil2019camera} embeddings to camera extrinsics.
            \item \textit{\uniDrive} \cite{li2025unidrive}: Transforming unseen ego \variation (target) images to car height (source) assuming objects at fixed distance parameter ($50m$) and then passing to the \monoThreeD model.
            \item \textit{\uniDrivePlus} \cite{li2025unidrive}: \uniDrive with distance parameter optimized per dataset.
        \item \textit{Oracle}: We also report the \textit{Oracle} \monoThreeD model, which is trained and tested on the \textbf{same} ego \variation.
        The Oracle serves as the \textbf{upper bound} of all baselines.
        \end{itemize}

    \subsection{\carla Error Analysis}
        We first report the error analysis of the baseline \gupNet~\cite{lu2021geometry} in \cref{tab:error_analysis} by replacing the predicted box data with the oracle parameters of the box as in~\cite{ma2021delving,kumar2024seabird}.
        We consider the GT box to be an oracle box for predicted box if the euclidean distance is less than $4m$ \cite{kumar2024seabird}.
        In case of multiple GT being matched to one box, we consider the oracle with the minimum distance.
        \cref{tab:error_analysis} shows that depth is the biggest source of error for \monoThreeD under ego \variation changes as also observed for single height settings in \cite{ma2021delving,kumar2022deviant,kumar2024seabird}.
        Note that the Oracle does not get $100\%$ results since we only replace box parameters in the baseline and consequently, the missed boxes in the baseline are not added.

        \begin{table*}[!t]
            \centering
            \scalebox{\scaleFraction}{
            \setlength\tabcolsep{0.23cm}
            \begin{tabular}{l l m dcd m dcd m dcd }
                \addlinespace[0.01cm]
                    \multirow{2}{*}{\threeD Detector} & \multirow{2}{*}{Method $\downarrowRHDSmall$ / $\egoHeightChange~(m)\rightarrowRHDSmall$} &  \multicolumn{3}{cm}{\apThreeDSeventy~\bracketPercentage~(\uparrowRHDSmall)} & \multicolumn{3}{cm}{\apThreeDFifty~\bracketPercentage~(\uparrowRHDSmall)} & \multicolumn{3}{c}{\MDE $(m)~[\approx 0]$}\\
                & & $-0.70$ & $0$ & $+0.76$ & $-0.70$ & $0$ & $+0.76$ & $-0.70$ & $0$ & $+0.76$\\
                \myTopRule
                \multirow{5}{*}{\gupNet \cite{lu2021geometry}} & Source & $10.13$ & \first{49.82} & $5.28$ & $47.15$  & \first{73.49} & $42.70$ & $+0.40$ & $+0.01$ & $-0.65$\\
                & \uniDrive \cite{li2025unidrive} & $10.05$ & \first{49.82} & $6.15$ & $47.15$  & \first{73.49} & $43.89$ & $+0.35$ & $+0.01$ & $-0.62$\\
                & \uniDrivePlus \cite{li2025unidrive} & $9.37$ & \first{49.82} & $13.00$ & $52.95$  & \first{73.49} & $55.57$ & $+0.31$ & $+0.01$ & $-0.46$\\
                & \cellcolor{methodColor}\textbf{\methodName} & \cellcolor{methodColor}\first{16.62} & \cellcolor{methodColor}$46.13$ & \cellcolor{methodColor}\first{24.50} & \cellcolor{methodColor}\first{57.00} & \cellcolor{methodColor}$67.83$ & \cellcolor{methodColor}\first{60.86} & \cellcolor{methodColor}$-0.15$ & \cellcolor{methodColor}$+0.00$ & \cellcolor{methodColor}$+0.07$ \\
                & Oracle & $70.25$ & $49.82$ & $62.93$  & $83.49$ & $73.49$ & $84.07$ & $-0.01$ & $+0.05$ & $+0.07$ \\
                \hline
                \multirow{5}{*}{\deviant \cite{kumar2022deviant}} & Source & $8.83$ & \first{49.88} & $4.43$ & $42.10$  & $72.79$ & $38.42$ & $+0.40$ & $+0.01$ & $-0.69$\\
                & \uniDrive \cite{li2025unidrive} & $8.21$ & $49.87$ & $3.75$ & $42.21$  & $72.79$ & $38.38$ & $+0.40$ & $+0.01$ & $-0.70$\\
                & \uniDrivePlus\cite{li2025unidrive} & $6.01$ & $49.87$ & $12.03$ & $43.99$  & $72.79$ & $50.67$ & $+0.38$ & $+0.01$ & $-0.50$\\
                & \cellcolor{methodColor}\textbf{\methodName} & \cellcolor{methodColor}\first{14.96} & \cellcolor{methodColor}$49.13$ & \cellcolor{methodColor}\first{23.66} & \cellcolor{methodColor}\first{52.68} & \cellcolor{methodColor}\first{72.95} & \cellcolor{methodColor}\first{60.98} & \cellcolor{methodColor}$-0.07$ & \cellcolor{methodColor}$+0.05$ & \cellcolor{methodColor}$+0.02$ \\
                & Oracle & $68.35$ & $49.88$ & $58.49$ & $84.03$  & $72.79$ & $83.42$ & $-0.04$ & $+0.01$ & $-0.1$\\
            \end{tabular}
            }
            \caption{\textbf{\carla \val Results with \resNetEighteen as the backbone of two 3D detectors}.
            \methodName \textbf{outperforms} all baselines, especially at bigger \hl{unseen ego \variations}.
            All methods except Oracle are trained on car height and tested on bot to truck height data.
            [Key: \firstKey{Best}]
            }
            \label{tab:det_results_carla_val_resnet}
        \end{table*}

        \begin{table*}[!t]
            \centering
            \scalebox{\scaleFraction}{
            \setlength\tabcolsep{0.2cm}
            \begin{tabular}{l l m dcd m dcd m dcd }
                \addlinespace[0.01cm]
                    \multirow{2}{*}{Change} & \multirow{2}{*}{From \rightarrowRHDSmall To ~~~~/ $\egoHeightChange~(m)\rightarrowRHDSmall$} &  \multicolumn{3}{cm}{\apThreeDSeventy~\bracketPercentage~(\uparrowRHDSmall)} & \multicolumn{3}{cm}{\apThreeDFifty~\bracketPercentage~(\uparrowRHDSmall)} & \multicolumn{3}{c}{\MDE $(m)~[\approx 0]$}\\
                & & $-0.70$ & $0$ & $+0.76$ & $-0.70$ & $0$ & $+0.76$ & $-0.70$ & $0$ & $+0.76$\\
                \myTopRule
                \gupNet \cite{lu2021geometry} & \mathDash & $9.46$ & $53.82$ & $7.23$ & $41.66$  & $76.47$ & $40.97$ & $+0.53$ & $+0.05$ & $-0.63$\\
                \hline
                \multirow{4}{*}{Merge} & Regress+Ground \rightarrowRHDSmall~Regress & $9.46$ & $53.82$ & $7.23$ & $41.66$  & $76.47$ & $40.97$ & $+0.53$ & $+0.05$ & $-0.63$ \\
                & Regress+Ground \rightarrowRHDSmall~Ground & $0.98$ & $26.61$ & $5.39$ & $14.21$  & $51.97$ & $31.42$ & $-0.80$ & $-0.01$ & $+0.55$ \\
                & Within Model \rightarrowRHDSmall~Offline & $12.86$ & $47.66$ & $18.36$ & $49.86$  & $76.30$ & $54.38$ & $+0.24$ & $+0.02$ & $-0.28$ \\
                & Simple Avg \rightarrowRHDSmall~Learned Avg & $8.25$ & \first{56.49} & $9.53$ & $38.58$ & \first{76.82} & $43.13$ & $+0.56$ & $-0.03$ & $-0.62$\\
                \hline
                Ground & \relu \rightarrowRHDSmall~No \relu & $0.60$ & $52.94$ & $0.07$ & $15.66$  & $74.79$ & $4.50$ & $-1.09$ & $-0.01$ & $+1.34$\\
                \hline
                Formulation & Product \rightarrowRHDSmall~Sum & $3.28$ & $37.22$ & $12.79$ & $17.28$  & $63.88$ & $47.09$ & $+0.56$ & $+0.09$ & $-0.22$\\
                \hline
                \cellcolor{methodColor}\textbf{\methodName} & \cellcolor{methodColor}\mathDash & \cellcolor{methodColor}\first{19.45} & \cellcolor{methodColor}$55.68$ & \cellcolor{methodColor}\first{27.33} & \cellcolor{methodColor}\first{53.40} & \cellcolor{methodColor}$74.47$ & \cellcolor{methodColor}\first{61.98} & \cellcolor{methodColor}$-0.07$ & \cellcolor{methodColor}$+0.05$ & \cellcolor{methodColor}$+0.02$ \\
                \hline
                Oracle & \mathDash & $70.96$ & $53.82$ & $62.25$ & $83.88$ & $76.47$ & $83.96$ & $+0.03$ & $+0.03$ & $+0.03$\\
            \end{tabular}
            }
            \caption{\textbf{Ablation Studies} of \gupNet + \methodName on the \carla \val split on \hl{unseen ego \variations}.
            [Key: \firstKey{Best}]
            }
            \label{tab:ablation}
            \vspace{-0.3cm}
        \end{table*}

    \subsection{\carla \Variation Robustness Results}
        \cref{tab:det_results_carla_val} presents the \carla \val results, reporting the \textbf{median} model over three different seeds with the model being the final checkpoint as \cite{kumar2022deviant}.
        It compares baselines and our \methodName on all \monoThreeD models - \gupNet \cite{lu2021geometry}, and \deviant \cite{kumar2022deviant}. 
        Except for Oracle, all models are trained from car height data and tested on all ego heights.
        \cref{tab:det_results_carla_val} confirms that \methodName outperforms other baselines on all the  \monoThreeD models, and results in a better height robust detector.
        We also plot these \apThreeD numbers and depth errors visually in \cref{fig:det_results_carla_val} for intermediate height changes to confirm our observations.
        The \MDE comparison in \cref{fig:mde_results_carla_val_gup} also shows the trend of baselines, while \methodName cancels the opposite trends in extrapolation.

        \noIndentHeading{Oracle Biases.}
        We further note biases in the Oracle models at big changes in ego \variation.
        This agrees the observations of \cite{tzofi2023towards} in the \bev segmentation task.
        While higher \apThreeD for a higher \variation could be explained by fewer occlusions due a higher \variation, higher \apThreeD at lower camera \variation is not explained by this hypothesis.
        We leave the analysis of higher Oracle numbers for a future work.

        \noIndentHeading{Results on Other Backbone.}
            We next investigate whether the extrapolation behavior holds for other backbone configurations of the two \threeD detectors following \cite{kumar2022deviant}.
            So, we replace their backbones with \resNetEighteen. 
            \cref{tab:det_results_carla_val_resnet} illustrates that extrapolation shows up in other backbones and \methodName again outperforms all baselines.
            The biggest gains are in big camera height changes, which agrees with \cref{tab:det_results_carla_val} results.

    \subsection{Ablation Studies}\label{sec:ablation}
        \cref{tab:ablation} ablates the design choices of \gupNet + \methodName on  \carla \val split, with the experimental setup of \cref{sec:experiments}.

        \noIndentHeading{Depth Merge.}
            We first analyze the impact of averaging the two depth estimates.
            Merging both regressed and ground-based depth estimates is crucial for optimal performance.
            Relying solely on the regressed depth gives good \inDomain performance but bad \ood performance.
            Using only ground depth generalizes poorly in both \inDomain and \ood settings, which is why it is not used in modern \monoThreeD models.
            However, it has a contrasting extrapolation \MDE compared to regression models.
            While offline merging of depth estimates from regression-only and ground-only models also improves extrapolation, it is slower and lacks end-to-end training.
            We also experiment with changing the simple averaging of \methodName to learned averaging.
            Simple average of \methodName outperforms learned one in \outDomain test because the learned average overfits to train distribution.

        \noIndentHeading{\relu{}ed Ground.}
            \cref{sec:bottom_ground_depth} says that \relu activation applied to the ground depth ensures spatial continuity and improves model training stability.
            Removing the \relu leads to training instability and suboptimal extrapolation to camera \variation. (The training also collapses in some cases).

        \noIndentHeading{Formulation.}
            \methodName estimates the projected \threeD bottom center by using the projected \threeD center and the \twoD height prediction.
            \cref{eq:bottom_center} predicts a coefficient $\alpha$ to determine the precise bottom center location.
            Product means predicting $\alpha$ and then multiplying by $(\pixVCenter\!-\!\pixVCenterTwoD)$ to obtain the shift, while sum means directly predicting the shift $\alpha$.
            Replace this product formulation by the sum formulation of $\alpha$ confirms that
            the product is more effective than the sum.

\section{Conclusion}

    This paper highlights the understudied problem of \monoThreeD generalization to unseen ego \variations.
    We first systematically analyze the impact of camera height variations on state-of-the-art \monoThreeD models, identifying depth estimation as the primary factor affecting performance.
    We mathematically prove and also empirically observe consistent negative and positive \trends in regressed and ground-based object depth estimates, respectively, under camera height changes.
    This paper then takes a step towards generalization to unseen camera heights and proposes \methodName.
    \methodName averages both depth estimates within the model to mitigate these opposing \trends.
    \methodName significantly enhances the generalization of \monoThreeD models to unseen camera heights, achieving \sota performance on the \carla dataset.
    We hope that this initial step towards generalization will contribute to safer AVs.
    Future work involves extending this idea to more \monoThreeD models.

{
    \small
    \bibliographystyle{ieeenat_fullname}
    \bibliography{references}
}

\addtocontents{toc}{\protect\setcounter{tocdepth}{3}}
\clearpage
\maketitlesupplementary

\begingroup
\let\clearpage\relax
\hypersetup{linkcolor=blue}
\tableofcontents
\endgroup

\renewcommand{\thesection}{A\arabic{section}}
\setcounter{section}{0}

\section{Additional Details and Proof}\label{sec:additional_proof}
     We now add more details and proofs which we could not put in the main paper because of the space constraints.

    \subsection{Proof of Ground Depth \cref{lemma:1}}\label{sec:proof_lemma}
        We reproduce the proof from \cite{yang2023gedepth} with our notations for the sake of completeness of this work.
        \begin{proof}
            We first rewrite the pinhole projection \cref{eq.pinhole} as:
            \begin{align}
            \begin{bmatrix}
            \varX \\
            \varY \\
            \varZ \\
            \end{bmatrix}
            =\rotation^{-1}
            (\intrinsic^{-1}
            \begin{bmatrix}
            \pixU\\
            \pixV\\
            1 \\
            \end{bmatrix}
            \posZ - \extrinsicTrans).
            \label{eq.rewrite}
            \end{align}
            We now represent the ray shooting from the camera optical center through each pixel as $\ray(\pixU, \pixV, \posZ)$.
            Using the matrix $\bm{A}\!=\!(a_{ij})\!=\!\rotation^{-1}\intrinsic^{-1}$, and the vector $\bm{B}\!=\!(b_i)\!=\!-\rotation^{-1}\extrinsicTrans$, we define the parametric ray as:
            \begin{align}
            \ray(\pixU, \pixV, \posZ) :
            \begin{cases}
            \varX = (a_{11}u +a_{12}v + a_{13}) \posZ +  b_1  \\
            \varY = (a_{21}u +a_{22}v + a_{23}) \posZ +  b_2  \\
            \varZ = (a_{31}u +a_{32}v + a_{33}) \posZ +  b_3  \\
            \end{cases}
            \label{eq.ray}
            \end{align}
            Moreover, the ground at a distance $h$ can be described by a plane, which is determined by the point $(0, \egoHeight, 0)$ in the plane and the normal vector $\overrightarrow{n}=(0,1,0)$:
            \begin{align}
            \ray \boldsymbol{\cdot} \overrightarrow{n} = \egoHeight.
            \label{eq.plane}
            \end{align}
            Then, the ground depth is the intersection point between this ray and the ground plane.
            Combining \cref{eq.ray,eq.plane}, the ground depth $\posZ$ of the pixel $(\pixU, \pixV)$ is:
            \begin{align}
                (a_{21}u +a_{22}v + a_{23}) \posZ +  b_2  &= \egoHeight \nonumber \\
                \implies \posZ &= \frac{\egoHeight - b_{2}}{a_{21}u + a_{22}v + a_{23}}.
            \end{align}
        \end{proof}

    \subsection{Proof of \cref{lemma:2}}\label{sec:proof_simple}
        We next derive \cref{lemma:2} from \cref{lemma:1} as follows.
        \begin{proof}
        {
            \begin{align}
            \bm{A}\!=\!(a_{ij})&=\rotation^{-1}\intrinsic^{-1}\!=\!\bm{I}^{-1}
            \begin{bmatrix}
            \focal & 0 & \ppointU\\
            0 & \focal & \ppointV\\
            0 & 0 & 1\\
            \end{bmatrix}^{-1} \nonumber\\
            &=\bm{I} \begin{bmatrix}
            \frac{1}{\focal} & 0 & \frac{-\ppointU}{\focal}\\
            0 & \frac{1}{\focal} & \frac{-\ppointV}{\focal}\\
            0 & 0 & 1\\
            \end{bmatrix}, \nonumber
            \end{align}
            }
            \noindent with rotation matrix $\rotation$ is identity $\bm{I}$ for forward cameras. So, $a_{21}\!=\!0,a_{22}\!=\!\frac{1}{\focal},a_{23}\!=\!\frac{-\ppointV}{\focal}$. Substituting $a_{21},a_{22},a_{23}$ in \cref{eq:gd_depth}, we get \cref{eq:gd_depth_simple}.
        \end{proof}

            \begin{figure*}[!t]
                \centering
                \begin{subfigure}{.38\linewidth}
                  \centering
                  \includegraphics[width=\linewidth]{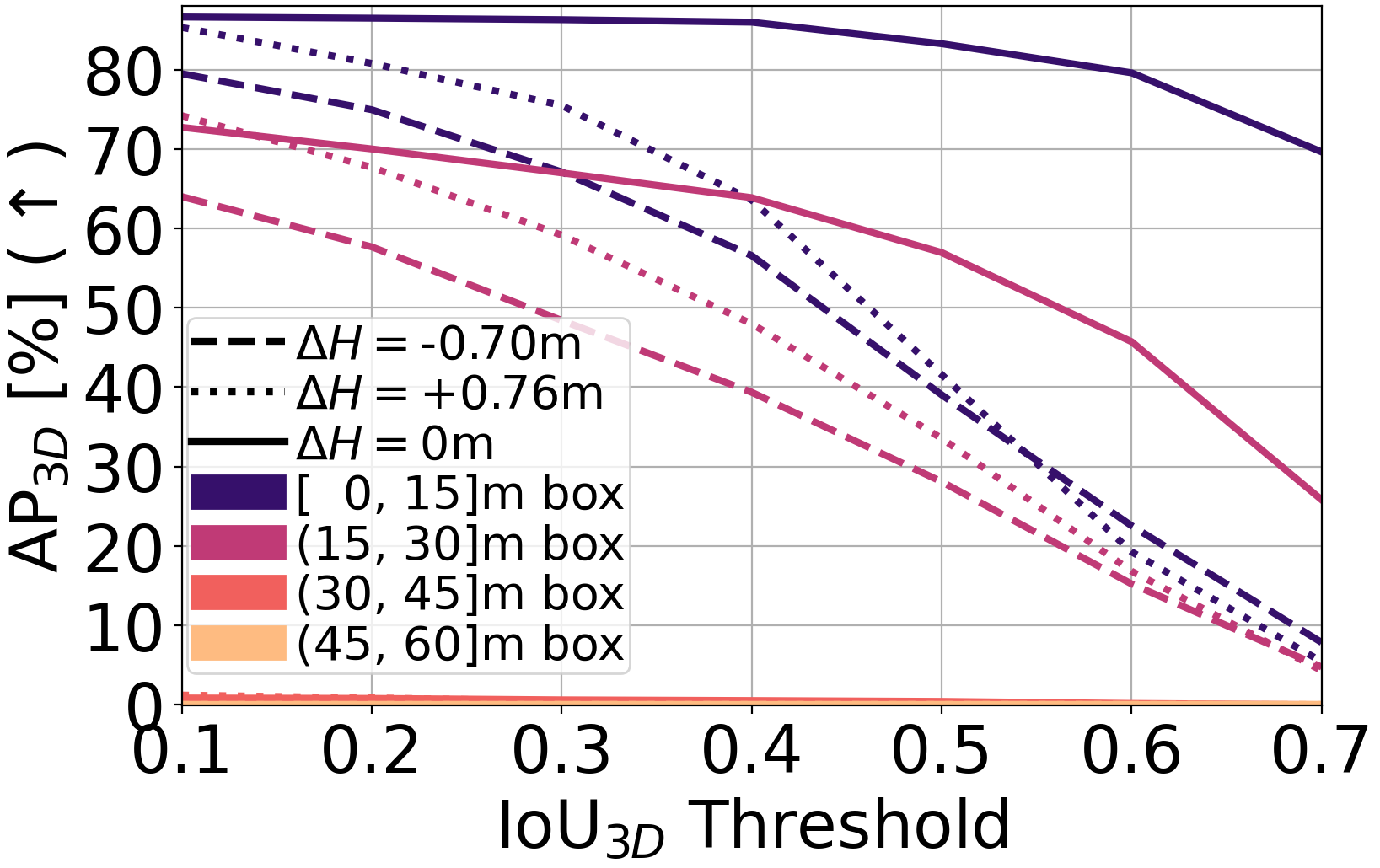}
                  \caption{\gupNet}
                \end{subfigure}~~~~
                \begin{subfigure}{.38\linewidth}
                  \centering
                  \includegraphics[width=\linewidth]{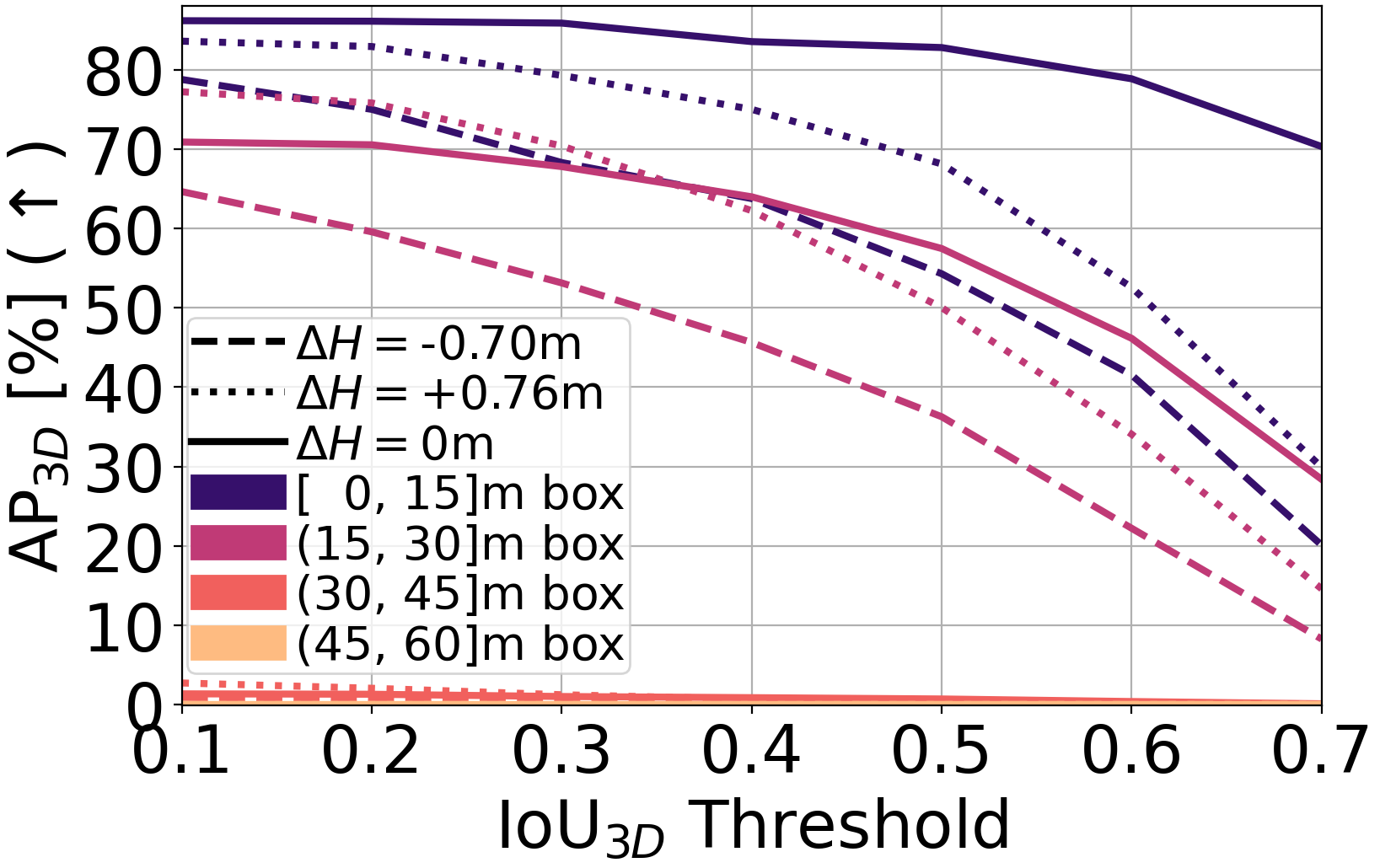}
                  \caption{\methodName}
                \end{subfigure}
                \caption{\textbf{\carla \val \apThreeD{} at different depths and \iouThreeD{} thresholds} with \gupNet. \methodName shows biggest gains on \iouThreeD $> 0.3$ for $[0, 30]m$ boxes.
                Note that $30m$ to $60m$ curves are present, but their performance is near zero, making them hard to see.
                }
                \label{fig:carla_ap_ground_truth_threshold}
            \end{figure*}

    \subsection{Pixel Shift with Ego Height Change.}\label{sec:pix_shift}
        We derive pixel shift with ego height change by backprojecting a pixel $\mathbf{p}\!=\!(\pixU,\pixV,\posZ)$ to \threeD, applying extrinsics change, and re-projecting to \twoD.
        The new point $\mathbf{p}'$ after height change of $\egoHeightChange$ is given by
        $\mathbf{p}'\!=\!\intrinsic[\rotation|\translation]\intrinsic^{-1}\mathbf{p}$ with usual notations.
        With height change inducing translation $\translation\!=\![0,\egoHeightChange,0]^T$ and not changing rotations ($\rotation\!=\!\identity$), $\mathbf{p}'\!=\!(\pixU,\pixV + \frac{f\egoHeightChange}{z},\posZ)$.

    \subsection{Extension to Camera Not Parallel to Ground}\label{sec:not_parallel}
        Following Sec. 3.3 of GEDepth \cite{yang2023gedepth}, we use the camera pitch $\delta$, and generalize \cref{eq:gd_depth} to obtain ground depth as
        \begin{align}
        \posZ &=\!\frac{\egoHeight\!-\!b_{2}\!\cos\delta\!-\!b_{3}\!\sin\delta}{[a_{21}\pixU\!+\!a_{22}\pixV\!+\!a_{23}]\cos\delta\!+\![a_{31}\pixU\!+\!a_{32}\pixV\!+\!a_{33}]\sin\delta} \nonumber \\
        &=\frac{H-b_{2}\cos\!\delta-b_{3}\!\sin\delta}{\frac{\pixV-\ppointV}{\focal}\cos\delta + \sin\delta}
        \label{eq:not_parallel}
        \end{align}.
        Note that if camera pitch $\delta=0$, this reduces to the usual form of \cref{eq:gd_depth} and \cref{eq:gd_depth_simple} respectively.
        Also, Th. 1 has a more general form with the pitch value, and remains valid for majority of the pitch angle ranges.

    \subsection{Extension to Not-flat Roads}\label{sec:not_flat}
        For non-flat roads, we assume that the road is made of multiple flat `pieces` of roads each with its own slope and we predict the slope of each pixel as in GEDepth \cite{yang2023gedepth}.
        To predict slope $\Hat{\delta}$ of each pixel, we first define a set of $N$ discrete slopes: $\{\tau_i,i\!=\!1,...,N\}$.
        We compute each pixel slope by linearly combining the discrete slopes with the predicted probability distribution $\{\Hat{p_i}\!\in\![0,1],\sum_{i}\!\Hat{p_i}\!=\!1\}$ over $N$ slopes $\Hat{\delta}\!=\!\sum_{i}\Hat{p_i}\tau_i$.
        We train the network to minimize the total loss:
        $L_{\text{total}}\!=\!L_{\text{det}}\!+\!\lambda_{\text{slope}} L_{\text{slope}}(\delta,\hat{\delta})$,
        where $L_{\text{det}}$ are the detection losses, and $L_{\text{slope}}$ is the slope classification loss.
        We next substitute the predicted slope in \cref{eq:not_parallel}.
        We do not run this experiment since planar ground is reasonable assumption for most driving scenarios within some distance.

    \subsection{\cref{theorem:1} in Slopy Ground}
        \cref{theorem:1} remains valid in slopy grounds. The key is the extrapolation behaviour of detectors, and not the ground depth itself.

        \begin{proof}
            Using \cref{eq:not_parallel}, the new depth $\newDepthGround$ is
            \begin{align}
                \newDepthGround &= \frac{\egoHeight + \egoHeightChange - b_{2}\cos\delta - b_{3}\sin\delta}{\frac{\pixVBottom + \frac{f\egoHeightChange}{z}\!-\!\ppointV}{\focal}\cos\delta\!+\!\sin\delta} \nonumber\\
                &\approx \frac{\egoHeight + \egoHeightChange - b_{2}\cos\delta - b_{3}\sin\delta}{\frac{\pixVBottom\!-\!\ppointV}{\focal}\cos\delta\!+\!\sin\delta} \nonumber \\
                &= \frac{\egoHeight - b_{2}\cos\delta - b_{3}\sin\delta}{\frac{\pixVBottom\!-\!\ppointV}{\focal}\cos\delta\!+\!\sin\delta} + \frac{\egoHeightChange}{~\frac{\pixVBottom\!-\!\ppointV}{\focal}\cos\delta\!+\!\sin\delta} \nonumber \\
                &= \oldDepthGround + \frac{\egoHeightChange}{~\frac{\pixVBottom\!-\!\ppointV}{\focal}\cos\delta\!+\!\sin\delta} \nonumber \\
                &\approx \depthGT + \noise + \dfrac{\focal\egoHeightChange}{(\pixVBottom\!-\!\ppointV)\cos\delta\!+\!\focal\sin\delta} \nonumber \\
                \implies \newDepthGround &- \depthGT \approx \noise + \dfrac{\focal\egoHeightChange}{(\pixVBottom\!-\!\ppointV)\cos\delta\!+\!\focal\sin\delta} \nonumber,
            \end{align}
            assuming the depth $\oldDepthGround$ at train height $\egoHeightChange\!=\!0$ is the GT depth $\depthGT$ added by a normal random variable $\noise(0, \normalVar)$ \cite{kumar2024seabird}.
            Taking expectation on both sides, mean depth error is
            \begin{align}
                \expect\Big(\newDepthGround - \depthGT\Big) &\approx
                \left(\dfrac{1}{{(\pixVBottom\!-\!\ppointV)\cos\delta\!+\!f\sin\delta}}\right)\focal\egoHeightChange.
            \end{align}
            \noindent
            Thus, this theorem remains valid for positive slopes and negative slopes $> -\arctan \left(\dfrac{\pixVBottom\!-\!\ppointV}{\focal}\right)$.
            The valid slope $\delta \in \left( -\arctan \left(\dfrac{\pixVBottom\!-\!\ppointV}{\focal}\right), \frac{\pi}{2}\right]$ radians.
            As an example, for bottom-most point $\ppointV = \frac{\heightImage}{2}$ and focal length $\focal = \frac{\heightImage}{2}$, the valid delta range $\delta \in \left(-\frac{\pi}{4}, \frac{\pi}{2}\right]$ radians.
            Since almost all real datasets have slopes $|\delta|\!<\!10$ degrees, the extrapolation behavior remains consistent as \cref{theorem:1}.
        \end{proof}

    \subsection{Unrealistic Assumptions}

        \noindent\textbf{Flat ground assumption does not hold in real-world datasets.}
            Real datasets like \kitti and \nuscenes are mostly flat-ground datasets.
            Recent methods such as MonoGround \cite{qin2022monoground} and Homography Loss \cite{gu2022homography} leverage this assumption.
            \methodName makes a crucial \textbf{first attempt to address the extrapolation problem} within this relevant and prevalent setting.

        \noindent\textbf{Over-idealized \cref{theorem:2}. Car has constant surface depth.}
            \cref{theorem:2} relies on Sec. 3.2 and Fig. 3 of \cite{dijk2019neural}, that proves that all depth estimators are regression models.
            The \monoThreeD task and previous works: \deviant~\cite{kumar2022deviant} (our baseline) and \monodetr predict depth at the car center, not its surface.
            Thus, \cref{theorem:2} addresses regression for this standard center-based depth prediction.

    \subsection{Error Trends For Far Objects}
        Note that the error trends of regression-based depth and ground-based depth do not completely cancel for far objects.
        However, the baseline \monoThreeD performance is already notably poor for far objects.
        \cref{fig:carla_ap_ground_truth_threshold} confirms that both regression-based baseline and \methodName performance are bad beyond $>30m$ range.

            \begin{figure*}[!t]
                \centering
                \begin{subfigure}{.32\linewidth}
                    \includegraphics[width=\linewidth]{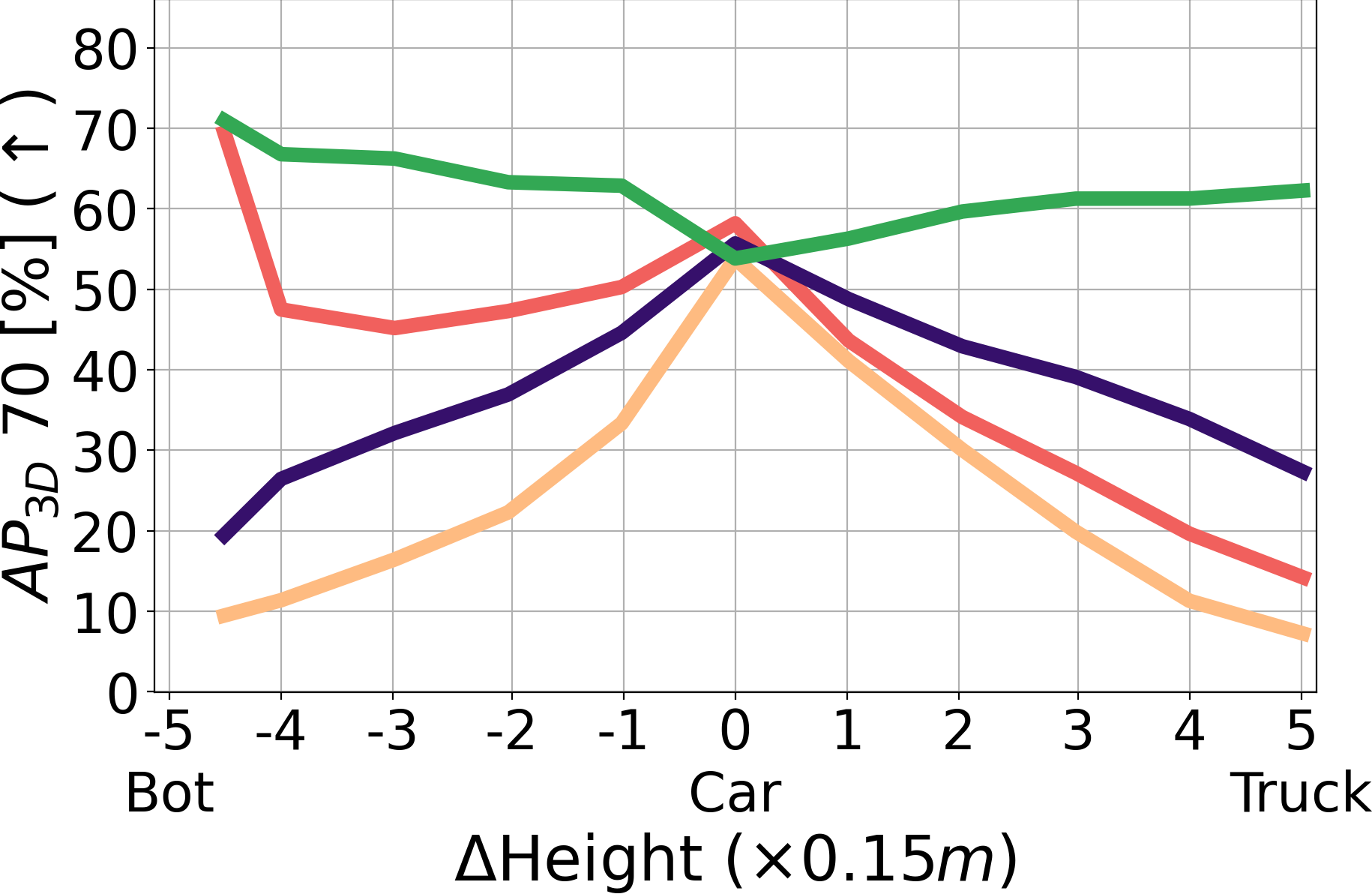}
                    \caption{\apThreeDSeventy \bracketPercentage{} comparison.}
                \end{subfigure}%
                \hfill
                \begin{subfigure}{.32\linewidth}
                    \includegraphics[width=\linewidth]{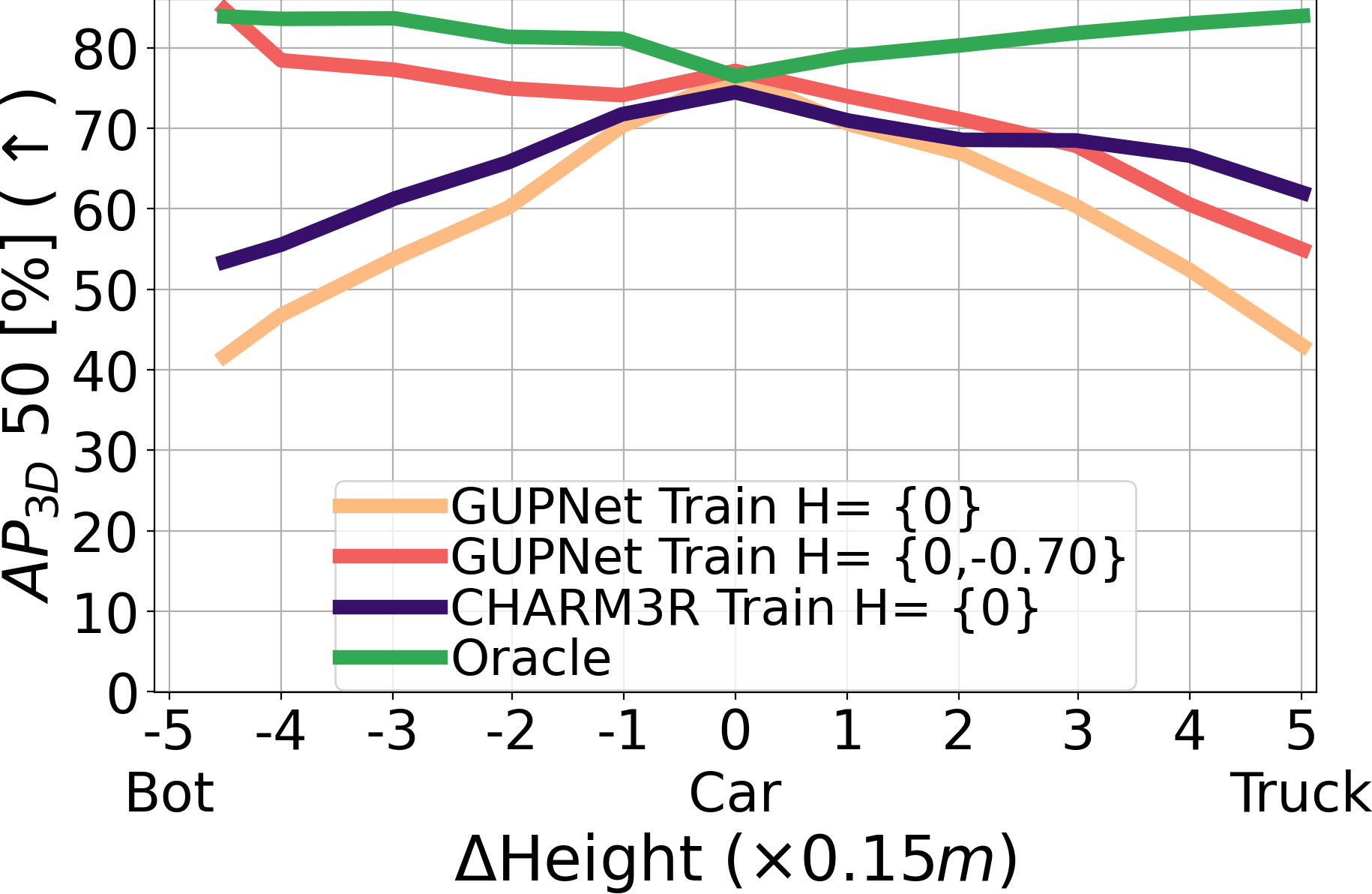}
                    \caption{\apThreeDFifty \bracketPercentage{} comparison.}
                \end{subfigure}
                \hfill
                \begin{subfigure}{.32\linewidth}
                    \includegraphics[width=\linewidth]{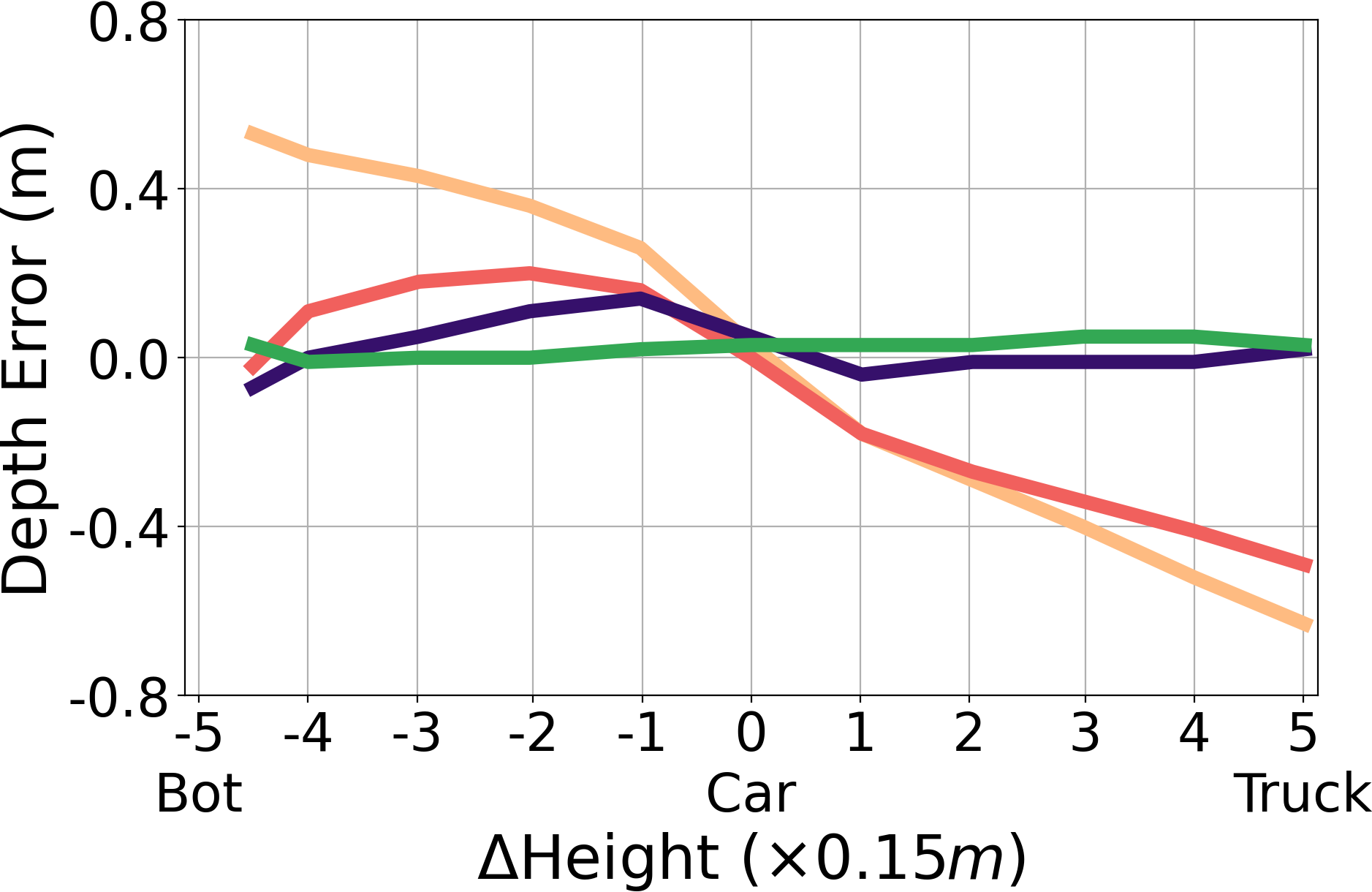}
                    \caption{\MDE comparison.}
                \end{subfigure}
                \caption{\textbf{\carla \val Results with \gupNet} detector after augmentation of \cite{tzofi2023towards}.
                Training a detector with both $\egoHeightChange=-0.70m$ and $\egoHeightChange=0m$ images produces better results at $\egoHeightChange=-0.70m$ and $\egoHeightChange=0m$, but \textbf{fails at unseen height images $\egoHeightChange=+0.76m$}.
                \methodName \textbf{outperforms} all baselines, especially at unseen bigger height changes.
                All methods except Oracle are trained on car height and tested on all heights.
                }
                \label{fig:det_results_carla_val_aug_more}
            \end{figure*}

\section{Additional Experiments }\label{sec:additional_expt}

        \begin{figure*}[!t]
            \centering
            \begin{subfigure}{.32\linewidth}
                \includegraphics[width=\linewidth]{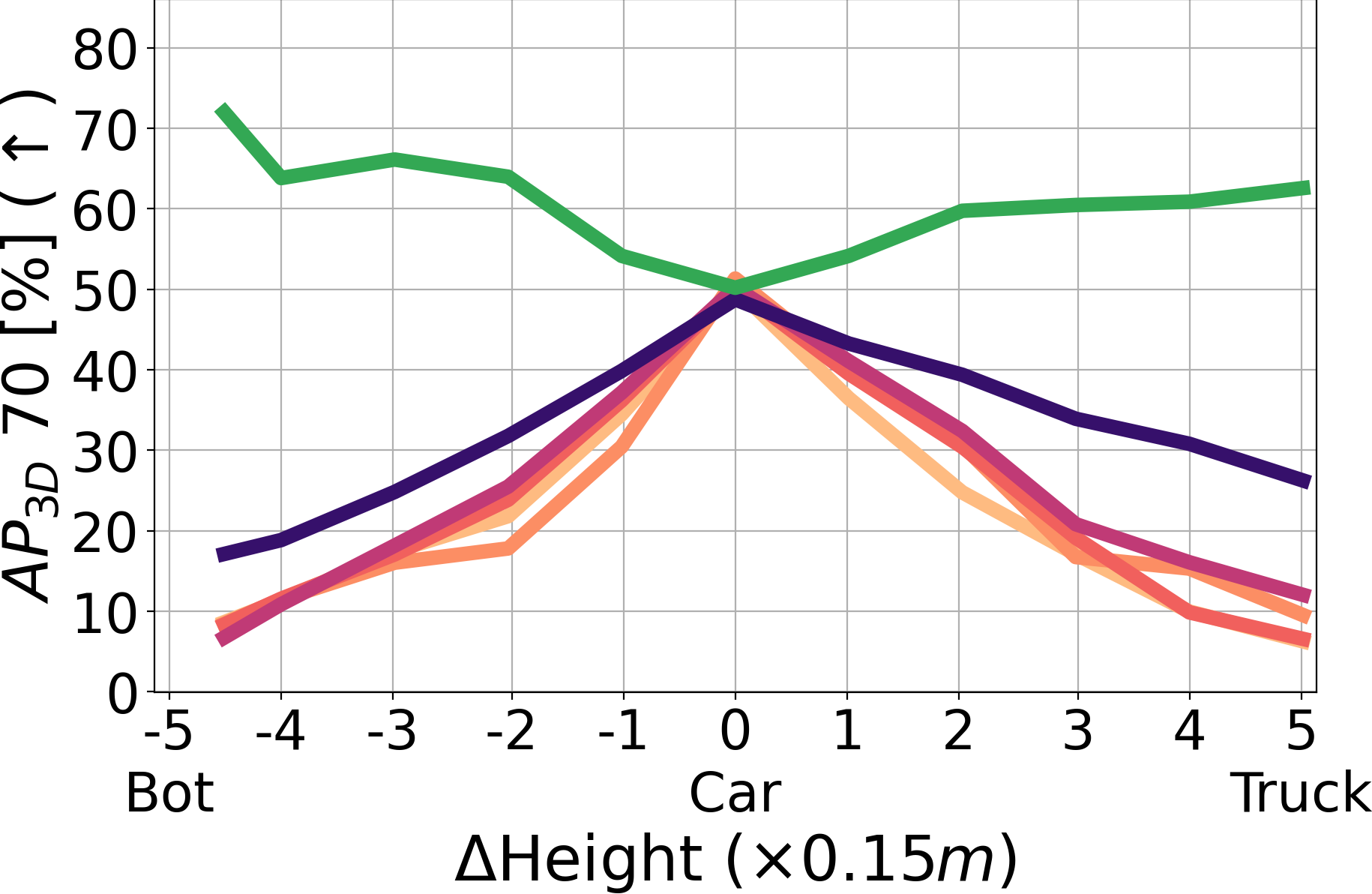}
                \caption{\apThreeDSeventy \bracketPercentage{} comparison.}
            \end{subfigure}%
            \hfill
            \begin{subfigure}{.32\linewidth}
                \includegraphics[width=\linewidth]{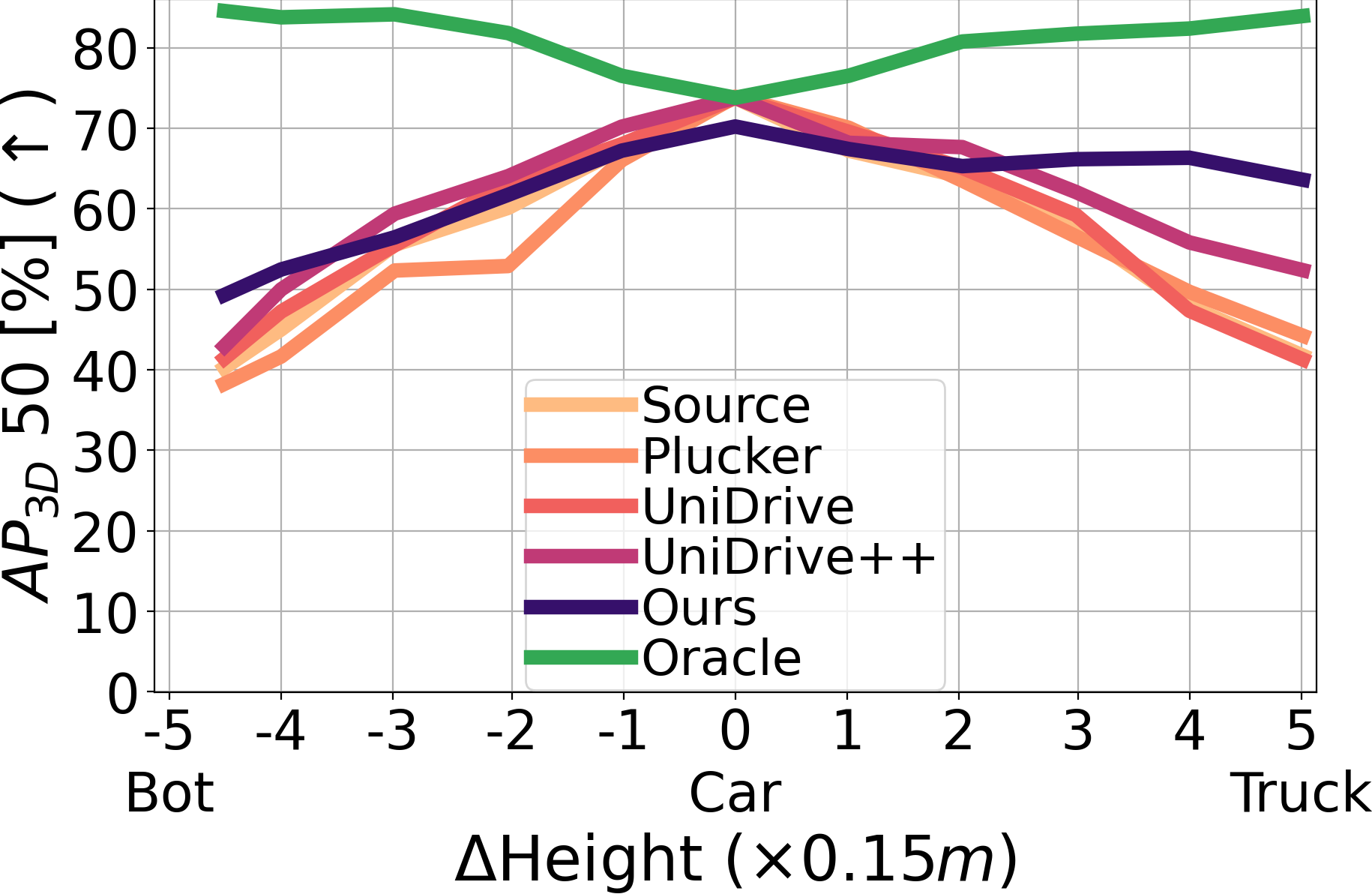}
                \caption{\apThreeDFifty \bracketPercentage{} comparison.}
            \end{subfigure}
            \hfill
            \begin{subfigure}{.32\linewidth}
                \includegraphics[width=\linewidth]{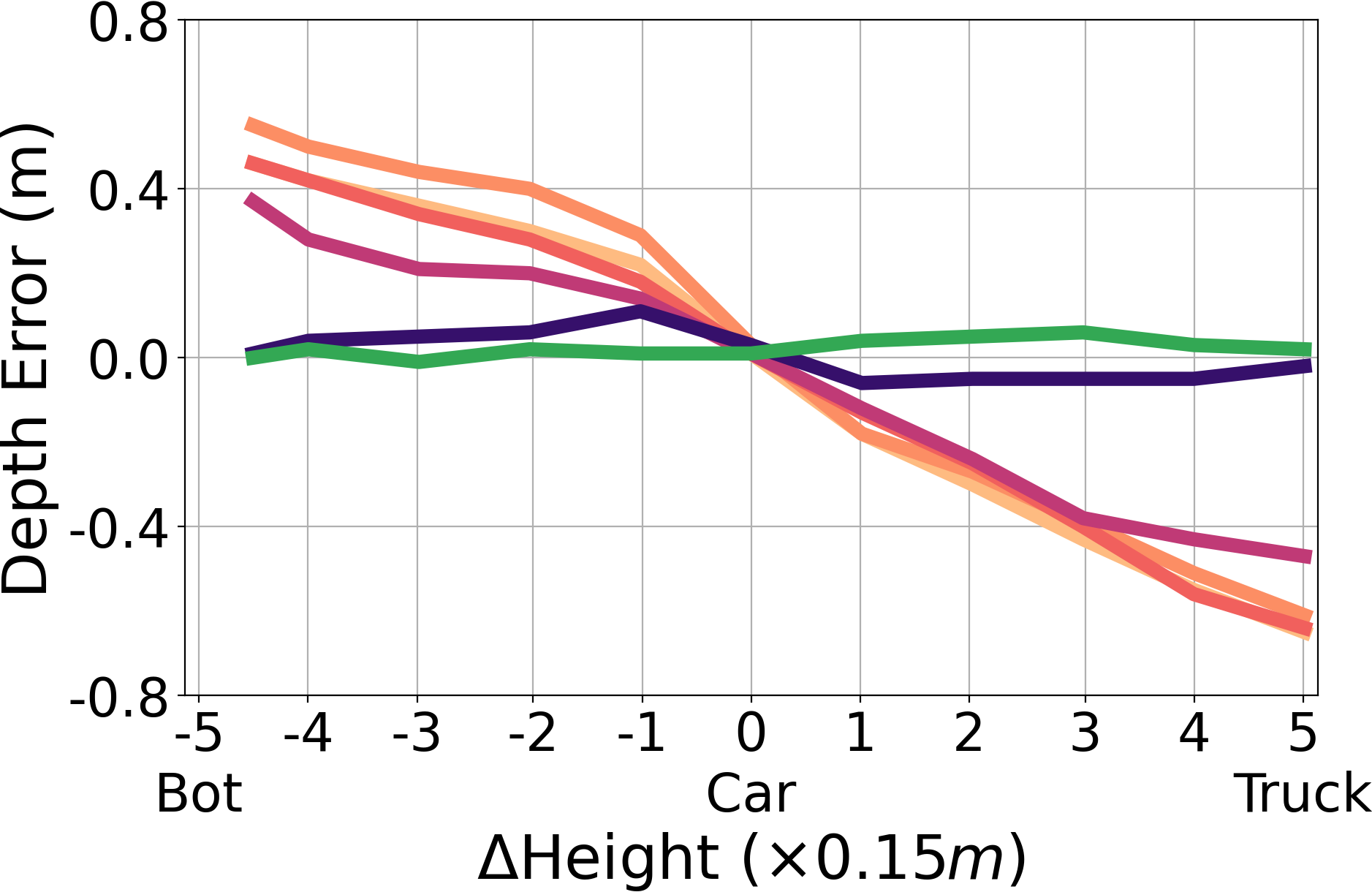}
                \caption{\MDE comparison.}
            \end{subfigure}
            \caption{\textbf{\carla \val Results with \deviant} detector.
            \methodName \textbf{outperforms} all baselines, especially at bigger height changes.
            All methods except Oracle are trained on car height and tested on all heights.
            Results of inference on height changes of $-0.70,0$ and $0.76$ meters are in \cref{tab:det_results_carla_val}.
            }
            \label{fig:det_results_carla_val_more}
        \end{figure*}

    We now provide additional details and results of the experiments evaluating \methodName{}’s performance.

    \noIndentHeading{Loss Function.}
        \methodName uses the same losses as baselines.
        \methodName's final depth estimate, being a fusion of regression and ground-based depth, does not need loss changes.

    \subsection{\carla \val Results}
        We first analyze the results on the synthetic \carla dataset further.

        \noIndentHeading{AP at different distances and thresholds.}
            We next compare the \apThreeD of the baseline \gupNet and \methodName in \cref{fig:carla_ap_ground_truth_threshold} at different distances in meters and \iouThreeD matching criteria of $0.1-0.7$ as in \cite{kumar2022deviant}.
            \cref{fig:carla_ap_ground_truth_threshold} shows that \methodName is effective over \gupNet at all depths and higher \iouThreeD thresholds.
            \methodName shows biggest gains on \iouThreeD $> 0.3$ for $[0, 30]m$ boxes.

        \noIndentHeading{Comparison with Augmentation-Methods.}
            \cref{sec:intro} of the paper says that the augmentation strategy falls short when the target height is \ood.
            We show this in \cref{fig:det_results_carla_val_aug_more}.
            Since authors of \cite{tzofi2023towards} do not release the NVS code, we use the ground truth images from height change $\egoHeightChange=-0.70m$ in training.
            \cref{fig:det_results_carla_val_aug_more} confirms that augmentation also improves the performance on $\egoHeightChange=-0.70m$ and $\egoHeightChange=0m$, but again falls short on unseen ego \variations $\egoHeightChange=+0.76m$.
            On the other hand, \methodName (even though trained on $\egoHeightChange=-0.70m$) outperforms such augmentation strategy at unseen ego heights $\egoHeightChange=+0.76m$.
            This shows the complementary nature of \methodName over augmentation strategies.

        \noIndentHeading{Reproducibility.}
            We ensure reproducibility of our results by repeating our experiments for 3 random seeds.
            We choose the final epoch as our checkpoint in all our experiments as \cite{kumar2022deviant,kumar2024seabird}.
            \cref{tab:det_results_carla_val_reproduce} shows the results with these seeds.
            \methodName outperforms the baseline \gupNet in both median and average cases.

        \begin{table*}[!t]
            \centering
            \scalebox{\scaleFraction}{
            \setlength\tabcolsep{0.23cm}
            \begin{tabular}{l l m dcd m dcd m dcdc}
                \addlinespace[0.01cm]
                \multirow{2}{*}{\threeD Detector} & \multirow{2}{*}{Seed $\downarrowRHDSmall$ / $\egoHeightChange~(m)\rightarrowRHDSmall$} & \multicolumn{3}{cm}{\apThreeDSeventy \bracketPercentage~(\uparrowRHDSmall)} & \multicolumn{3}{cm}{\apThreeDFifty \bracketPercentage~(\uparrowRHDSmall)} & \multicolumn{3}{c}{\MDE $(m)~[\approx 0]$}\\
                &  & $-0.70$ & $0$ & $+0.76$ & $-0.70$ & $0$ & $+0.76$ & $-0.70$ & $0$ & $+0.76$\\
                \myTopRule
                \multirow{4}{*}{\gupNet \cite{lu2021geometry}}
                & $111$  & $12.24$ & $55.98$ & $7.53$ & $44.14$ & $76.37$ & $41.32$ & $+0.48$ & $+0.00$ & $-0.64$\\
                & $444$  & $9.46$ & $53.82$ & $7.23$ & $41.66$ & $76.47$ & $40.97$ & $+0.53$ & $+0.03$ & $-0.63$\\
                & $222$  & $10.35$ & $52.94$ & $10.79$ & $41.67$ & $75.80$ & $46.45$ & $+0.53$ & $+0.01$ & $-0.57$\\
                \hhline{|~|-----------|}
                & Average  & $10.68$ & $54.25$ & $8.52$ & $42.49$ & \first{76.21} & $43.58$ & $+0.51$ & $+0.01$ & $-0.61$\\
                \myTopRule
                \multirow{4}{*}{+~\methodName}
                & $111$  & $19.99$ & $58.16$ & $29.96$ & $54.15$ & $74.10$ & $64.27$ & $+0.09$ & $+0.00$ & $-0.03$\\
                & $444$  & $19.45$ & $55.68$ & $27.33$ & $53.40$ & $74.47$ & $61.98$ & $+0.07$ & $+0.05$ & $-0.02$\\
                & $222$  & $17.41$ & $53.57$ & $27.77$ & $54.30$ & $74.83$ & $64.42$ & $+0.12$ & $+0.01$ & $-0.09$\\
                \hhline{|~|-----------|}
                & Average  & \first{18.95} & \first{55.80} & \first{28.35} & \first{53.95} & $74.47$ & \first{63.56} & $+0.09$ & $+0.02$ & $-0.05$\\
                \myTopRule
                Oracle & \mathDash & $70.96$ & $53.82$ & $62.25$ & $83.88$ & $76.47$ & $83.96$ & $+0.03$ & $+0.03$ & $+0.03$\\
            \end{tabular}
            }
            \caption{\textbf{Reproducibility Results.}
            \methodName \textbf{outperforms} all other baselines on \carla \val split, especially at bigger \hl{unseen ego \variations} in both median (Seed=$444$) and average cases.
            All except Oracle are trained on car height $\egoHeightChange=0m$ and tested on bot to truck height data.
            [Key: \firstKey{Best}]
            }
            \label{tab:det_results_carla_val_reproduce}
        \end{table*}

         \begin{table}[!t]
            \centering
            \scalebox{\scaleFraction}{
            \setlength\tabcolsep{0.1cm}
            \begin{tabular}{l c m c m ccc }
                \addlinespace[0.00cm]
                \textbf{\val} & \textbf{Ego Ht} $(m)$ & \textbf{\#Images} & \textbf{Car} $(k)$ & \textbf{Ped} $(k)$\\
                \myTopRule
                \nuscenes & $1.51$ & $6{,}019$ & $18$  & $7$ \\
                \coda & $0.75$ & $4{,}176$ & $4$ & $86$ \\
            \end{tabular}
            }
            \caption{\textbf{Dataset statistics.}
            \nuscenes \val has more Cars compared to Pedestrians, while \coda \val has more Pedestrians than Cars.
            }
            \label{tab:dataset_statistics}
        \end{table}

        \begin{table*}[!t]
            \centering
            \scalebox{\scaleFraction}{
            \setlength\tabcolsep{0.15cm}
            \begin{tabular}{l l m dc m dc }
                \addlinespace[0.00cm]
                \multirow{2}{*}{\threeD Detector} & \multirow{2}{*}{Method} &\multicolumn{2}{cm}{Car \apThreeDFifty~\bracketPercentage~(\uparrowRHDSmall)} & \multicolumn{2}{c}{Ped \apThreeDThirty~\bracketPercentage~(\uparrowRHDSmall)} \\
                & & \coda & \nuscenes & \coda & \nuscenes \\
                \myTopRule
                \multirow{5}{*}{\!\gupNet{} \cite{lu2021geometry}}
                & Source  & $0.02$ & \first{18.42} & $0.01$ &\first{2.93} \\
                & \uniDrive \cite{li2025unidrive} & $0.02$ & \first{18.42} & $0.01$ &\first{2.93}\\
                & \uniDrivePlus{}\!\cite{li2025unidrive} & \second{0.03} & \first{18.42} & \second{0.02} &\first{2.93}\\
                & \cellcolor{methodColor}\textbf{\methodName} & \cellcolor{methodColor}\first{0.30} & \cellcolor{methodColor}\second{14.80} & \cellcolor{methodColor}\first{0.05} & \cellcolor{methodColor}\second{1.26}\\
                \hhline{|~|-----|}
                & Oracle & $28.56$ & $18.42$ & $30.31$ & $2.93$ \\
            \end{tabular}
            }
            \caption{\textbf{\nuscenes to \coda \val Results.}
            \methodName \textbf{outperforms} all baselines, especially at \hl{unseen height changes}.
            [Key: \firstKey{Best}, \secondKey{Second Best}, Ped= Pedestrians]
            }
            \label{tab:det_results_nusc_to_coda}
        \end{table*}

        \begin{figure*}[!t]
            \centering
            \begin{subfigure}{.38\linewidth}
              \centering
              \includegraphics[width=\linewidth]{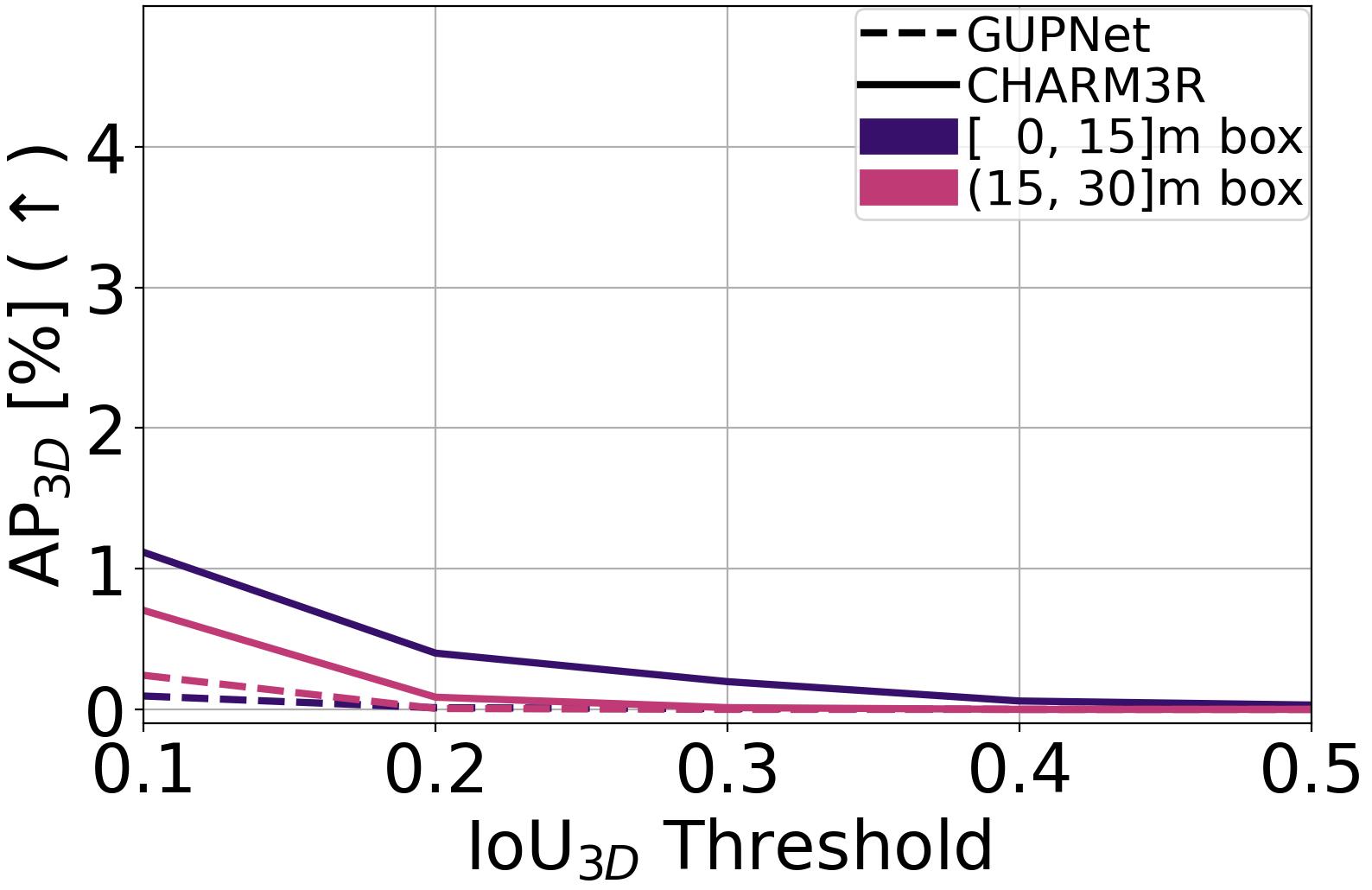}
              \caption{\coda Car}
            \end{subfigure}~~~~
            \begin{subfigure}{.38\linewidth}
              \centering
              \includegraphics[width=\linewidth]{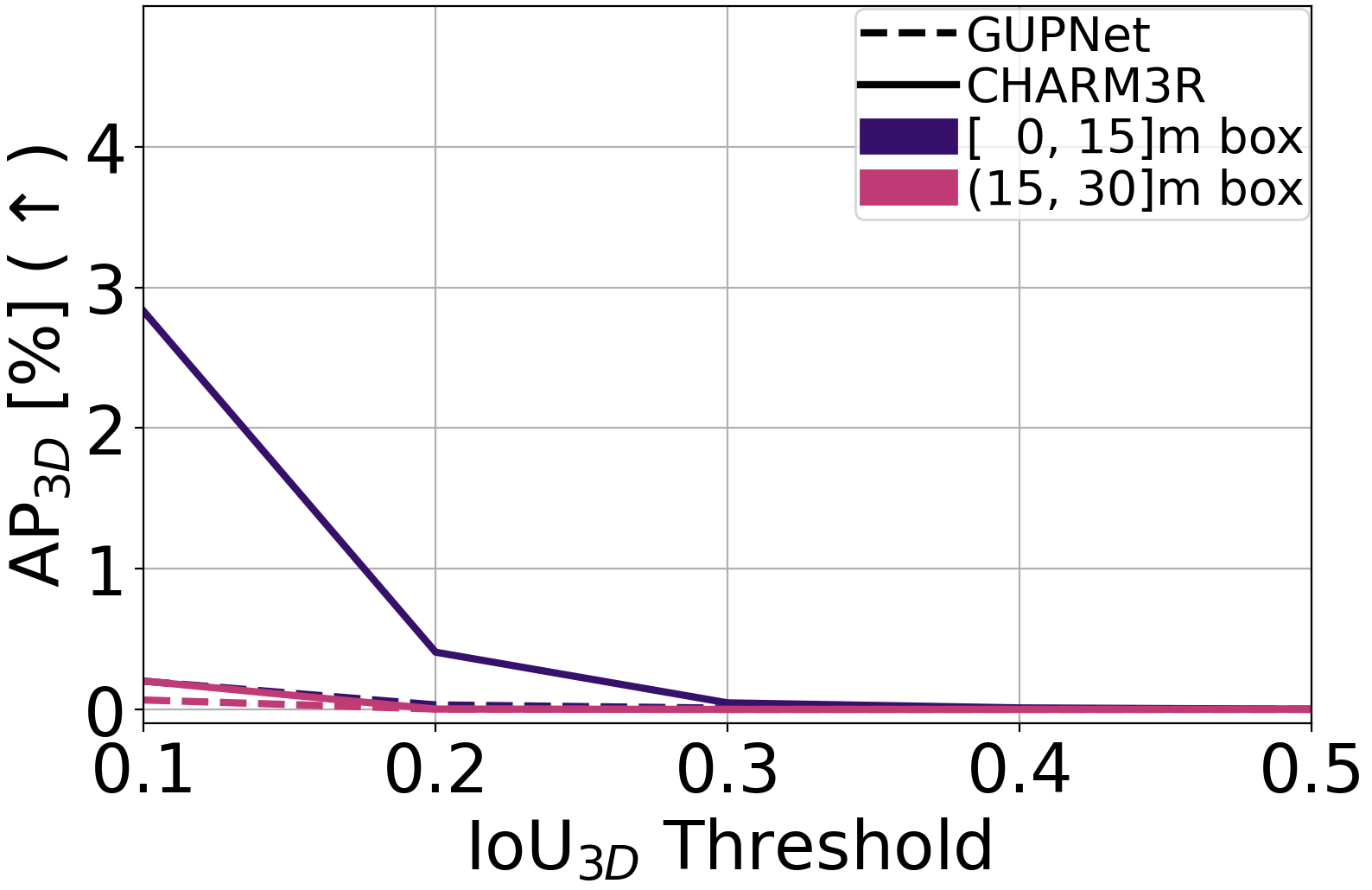}
              \caption{\coda Pedestrian}
            \end{subfigure}
            \caption{\textbf{\coda \val \apThreeD{} at different depths and \iouThreeD{} thresholds} with \gupNet trained on \nuscenes. \methodName shows biggest gains on \iouThreeD $< 0.3$ for $[0, 30]m$ boxes. }
            \label{fig:coda_ap_ground_truth_threshold}
        \end{figure*}

        \noIndentHeading{Results with \deviant.}
            We next additionally plot the robustness of \methodName with other methods on the \deviant detector \cite{kumar2022deviant} in \cref{fig:det_results_carla_val_more}
            The figure confirms that \methodName works even with \deviant and produces \sota robustness to unseen ego \variations.

    \subsection{\nuscenes $\rightarrowRHD$ \coda \val Results}
        To test our claims further in real-life, we use two real datasets: the \nuscenes dataset \cite{caesar2020nuscenes} and the recently released \coda \cite{zhang2023utcoda} datasets.
        \nuscenes has ego camera at height $1.51m$ above the ground, while the \coda is a robotics dataset with ego camera at a height of $0.75m$ above the ground.
        \cref{tab:dataset_statistics} shows the statistics of these two datasets.
        This experiment uses the following data split:
        \begin{itemize}
            \item \textit{\nuscenes \val Split.}
            This split \cite{caesar2020nuscenes} contains $28{,}130$ training and $6{,}019$ validation images from the front camera as \cite{kumar2022deviant}.
            \item \textit{\coda \val Split.}
            This split \cite{zhang2023utcoda} contains $19{,}511$ training and $4{,}176$ validation images. We only use this split for testing.
        \end{itemize}

        We train the \gupNet detector with $10$ \nuscenes classes and report the results with the \kitti metrics on both \nuscenes val and \coda \val splits.

        \noIndentHeading{Main Results.}
            We report the main results in \cref{tab:det_results_nusc_to_coda} paper.
            The results of \cref{tab:det_results_nusc_to_coda} shows gains on both Cars and Pedestrians classes of \coda val dataset.
            The performance is very low, which we believe is because of the domain gap between \nuscenes and \coda datasets.
            These results further confirm our observations that unlike \twoD detection, generalization across unseen datasets remains a big problem in the \monoThreeD task.

        \noIndentHeading{AP at different distances and thresholds.}
            To further analyze the performance, we next plot the \apThreeD of the baseline \gupNet and \methodName in \cref{fig:coda_ap_ground_truth_threshold} at different distances in meters and \iouThreeD matching criteria of $0.1-0.5$ as in \cite{kumar2022deviant}.
            \cref{fig:coda_ap_ground_truth_threshold} shows that \methodName is effective over \gupNet at all depths and lower \iouThreeD thresholds.
            \methodName shows biggest gains on \iouThreeD $< 0.3$ for $[0, 30]m$ boxes.
            The gains are more on the Pedestrian class on \coda since \coda captures UT Austin campus scenes, and therefore, has more pedestrians compared to cars.
            \nuscenes captures outdoor driving scenes in Boston and Singapore, and therefore, has more cars compared to pedestrians.
            We describe the statistics of these two datasets in \cref{tab:dataset_statistics}.

    \subsection{Qualitative Results.}

\begin{figure*}[!t]
            \centering
            \begin{subfigure}{0.4\linewidth}
                \includegraphics[width=\linewidth]{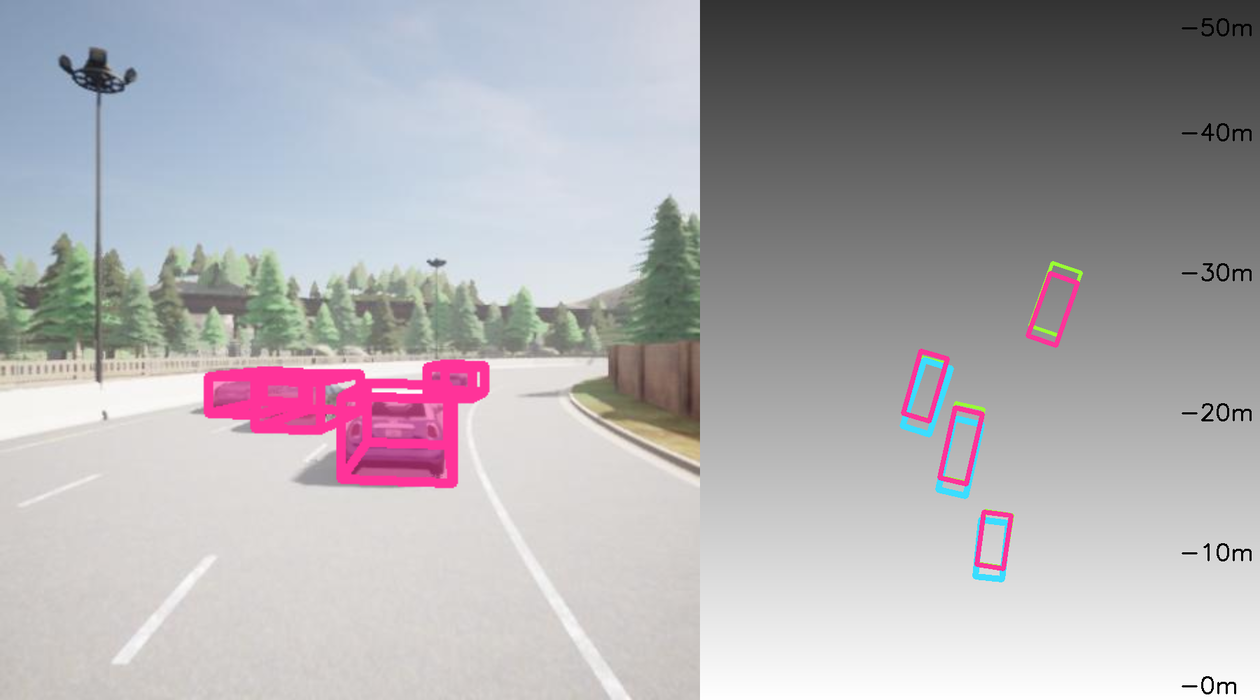}
            \end{subfigure}
            \begin{subfigure}{0.4\linewidth}
                \includegraphics[width=\linewidth]{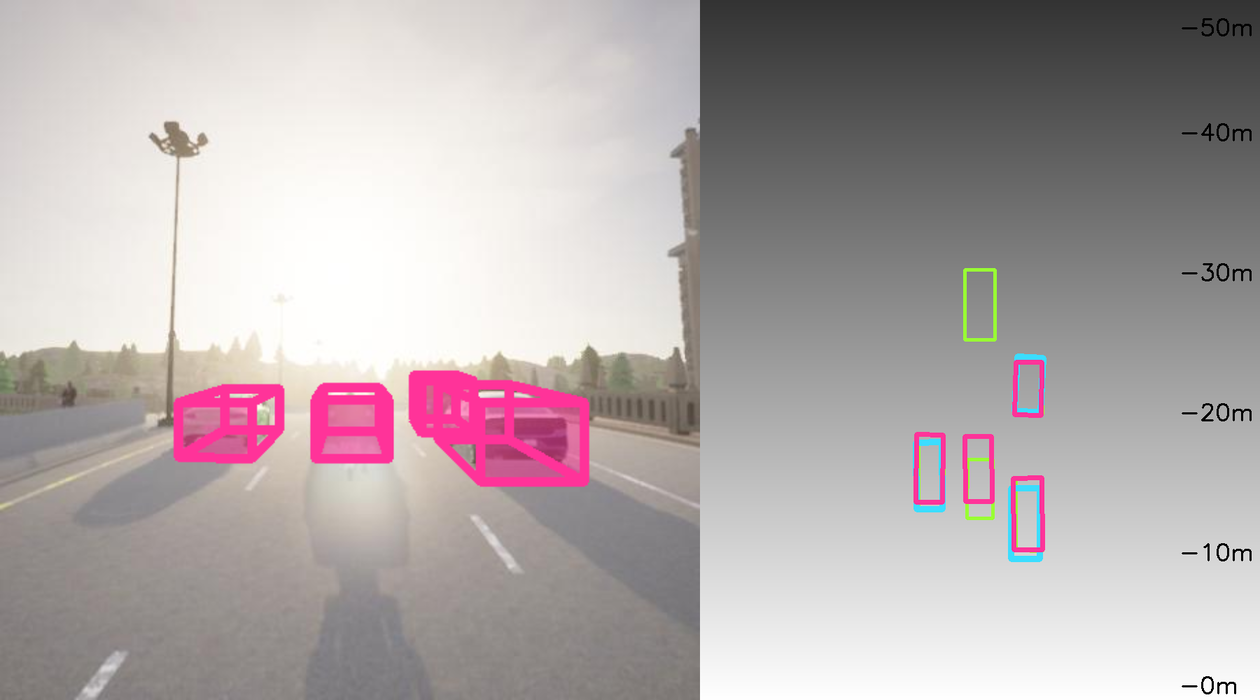}
            \end{subfigure}
            \begin{subfigure}{0.4\linewidth}
                \includegraphics[width=\linewidth]{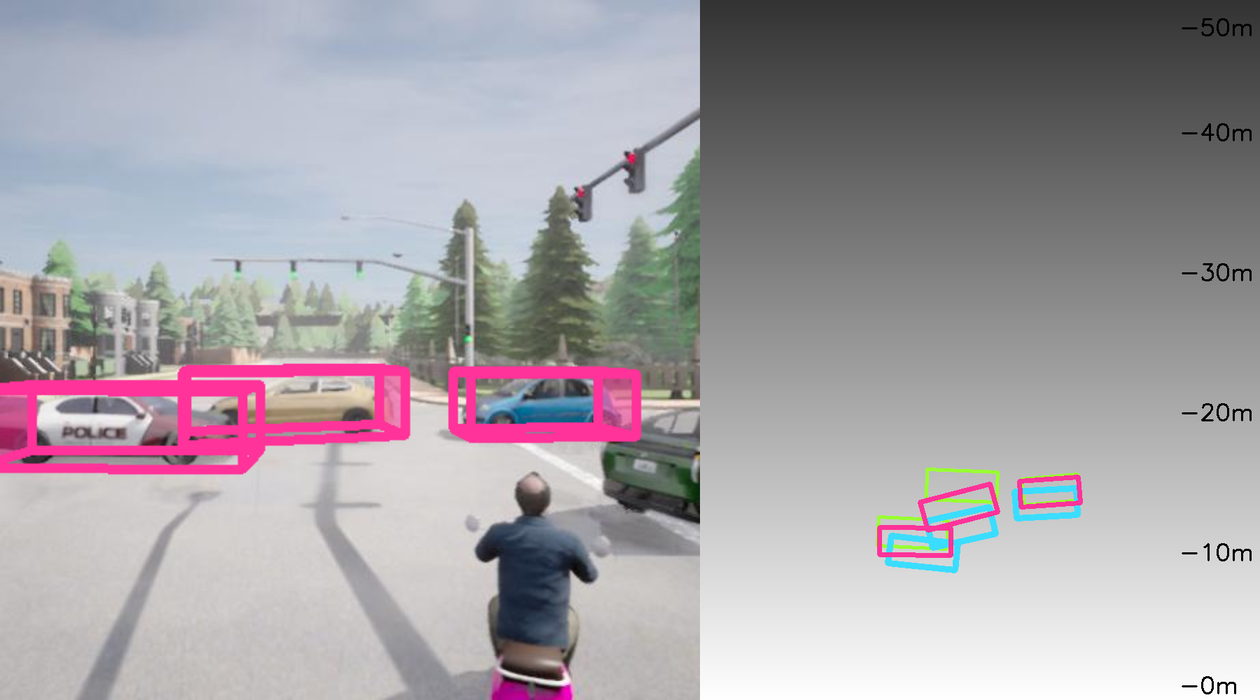}
            \end{subfigure}
            \begin{subfigure}{0.4\linewidth}
                \includegraphics[width=\linewidth]{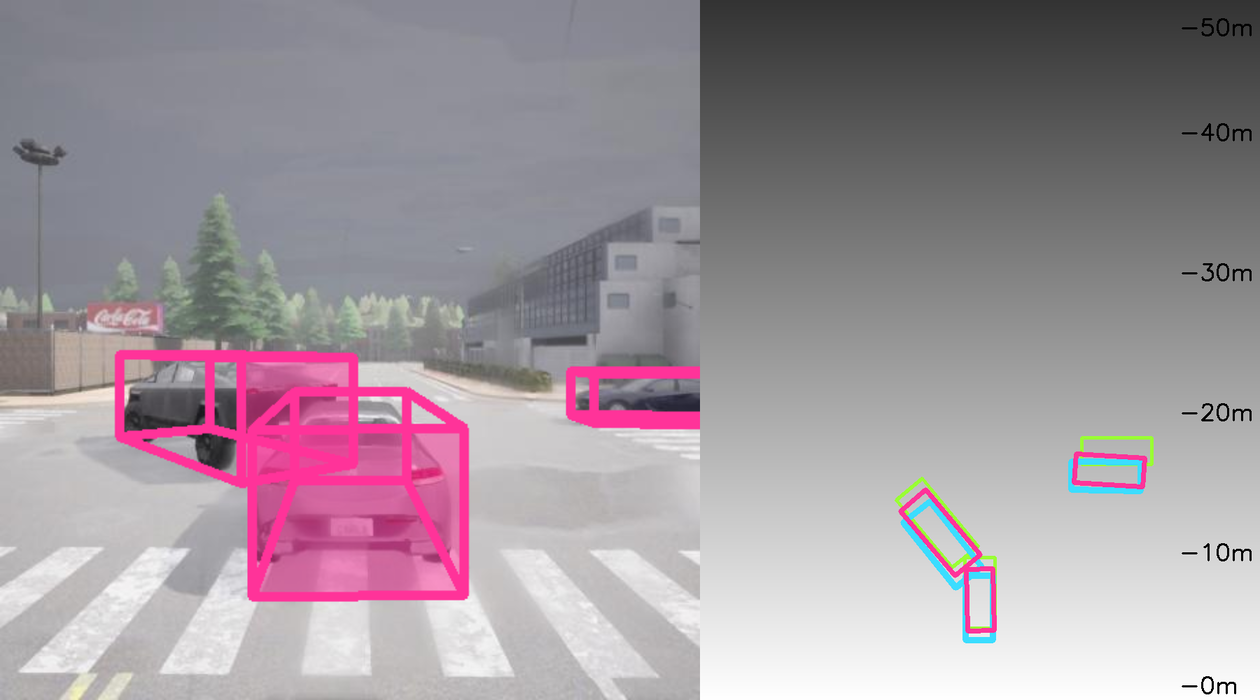}
            \end{subfigure}
            \begin{subfigure}{0.4\linewidth}
                \includegraphics[width=\linewidth]{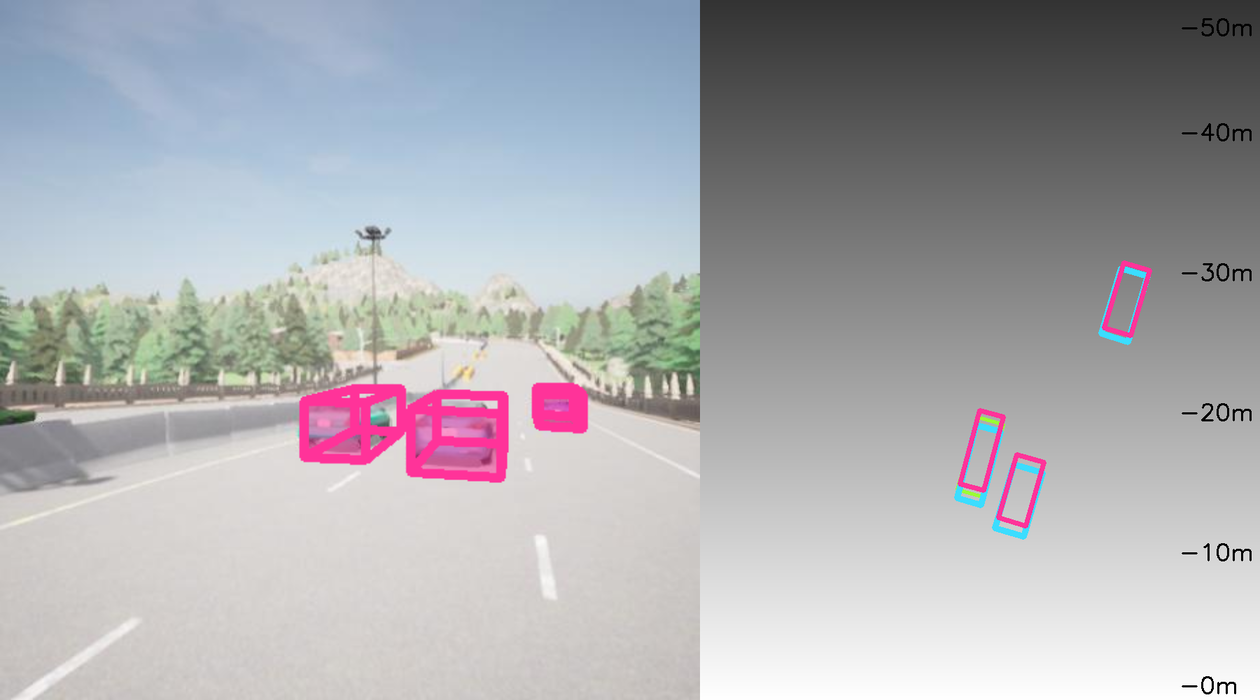}
            \end{subfigure}
            \begin{subfigure}{0.4\linewidth}
                \includegraphics[width=\linewidth]{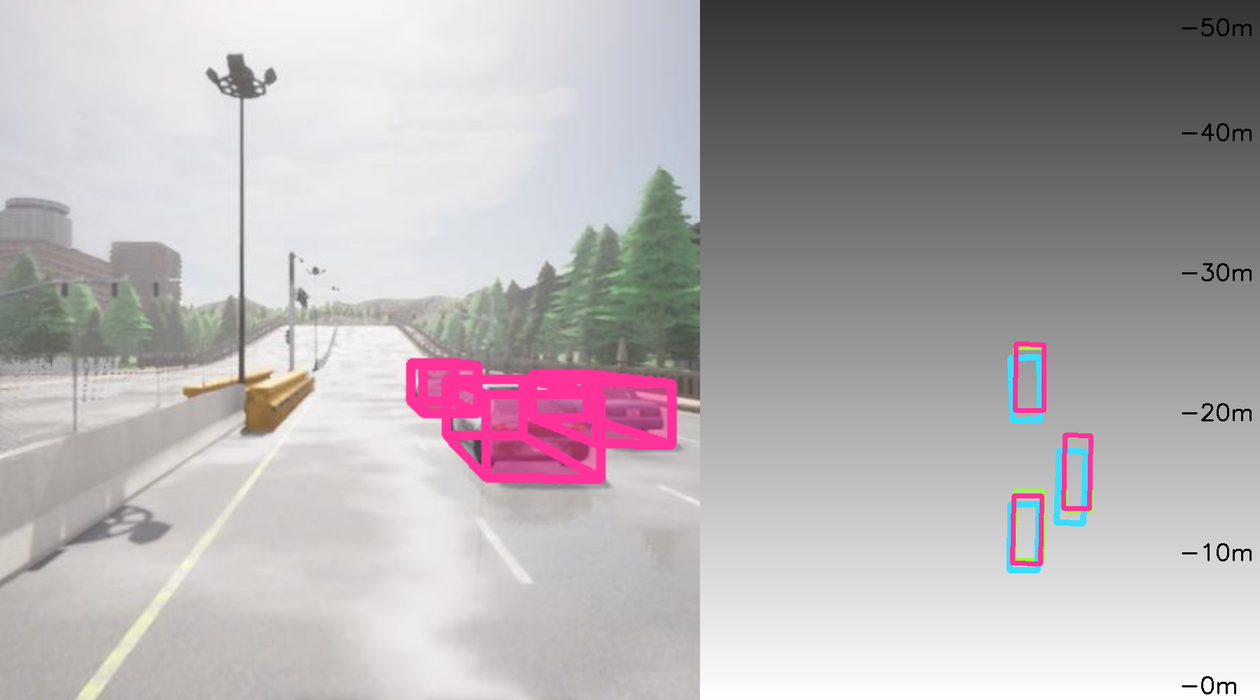}
            \end{subfigure}
            \begin{subfigure}{0.4\linewidth}
                \includegraphics[width=\linewidth]{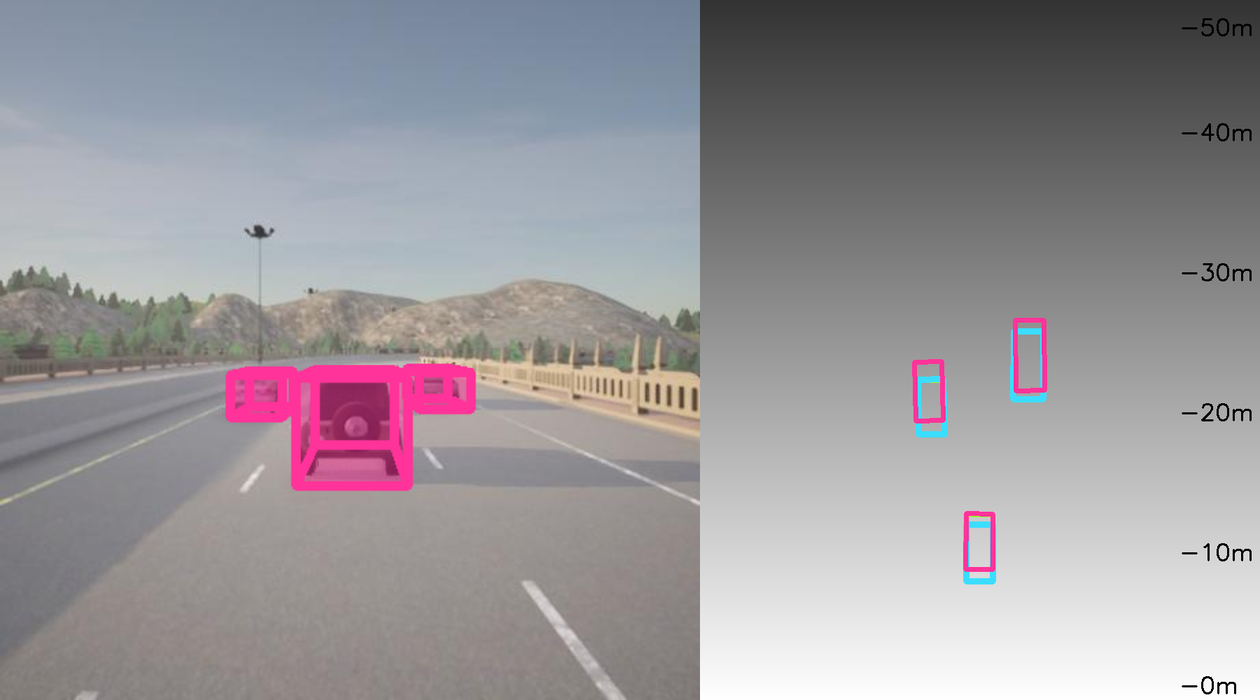}
            \end{subfigure}
            \begin{subfigure}{0.4\linewidth}
                \includegraphics[width=\linewidth]{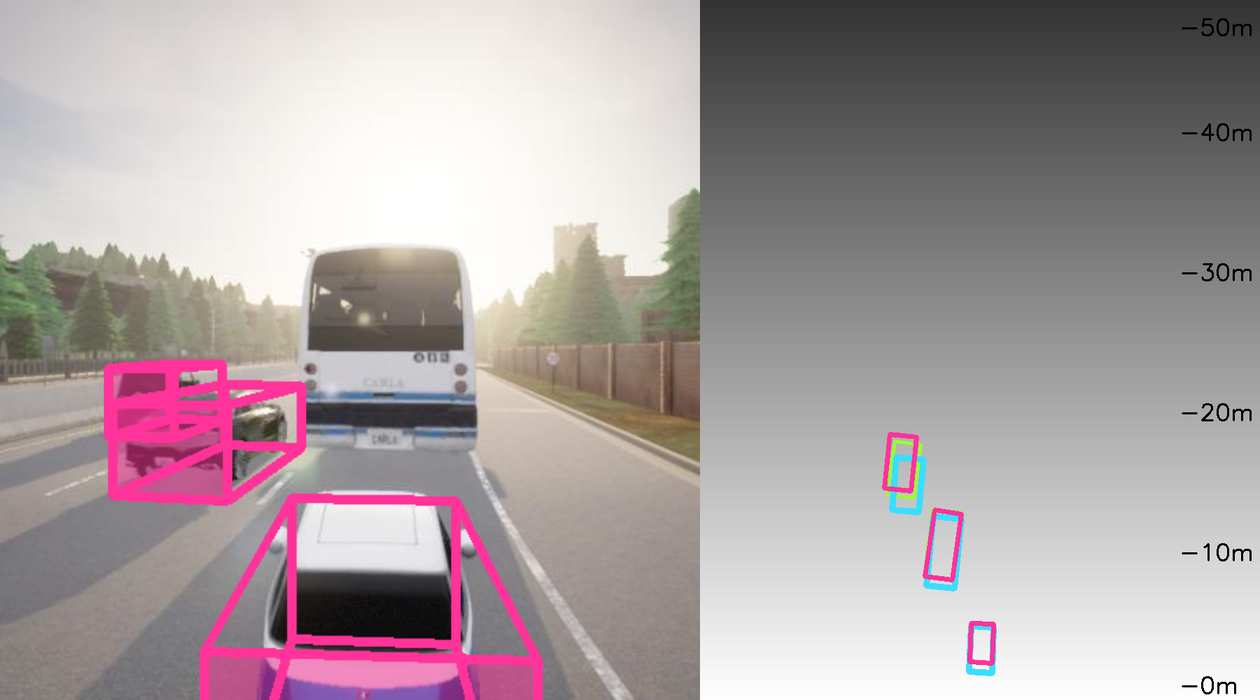}
            \end{subfigure}
            \begin{subfigure}{0.4\linewidth}
                \includegraphics[width=\linewidth]{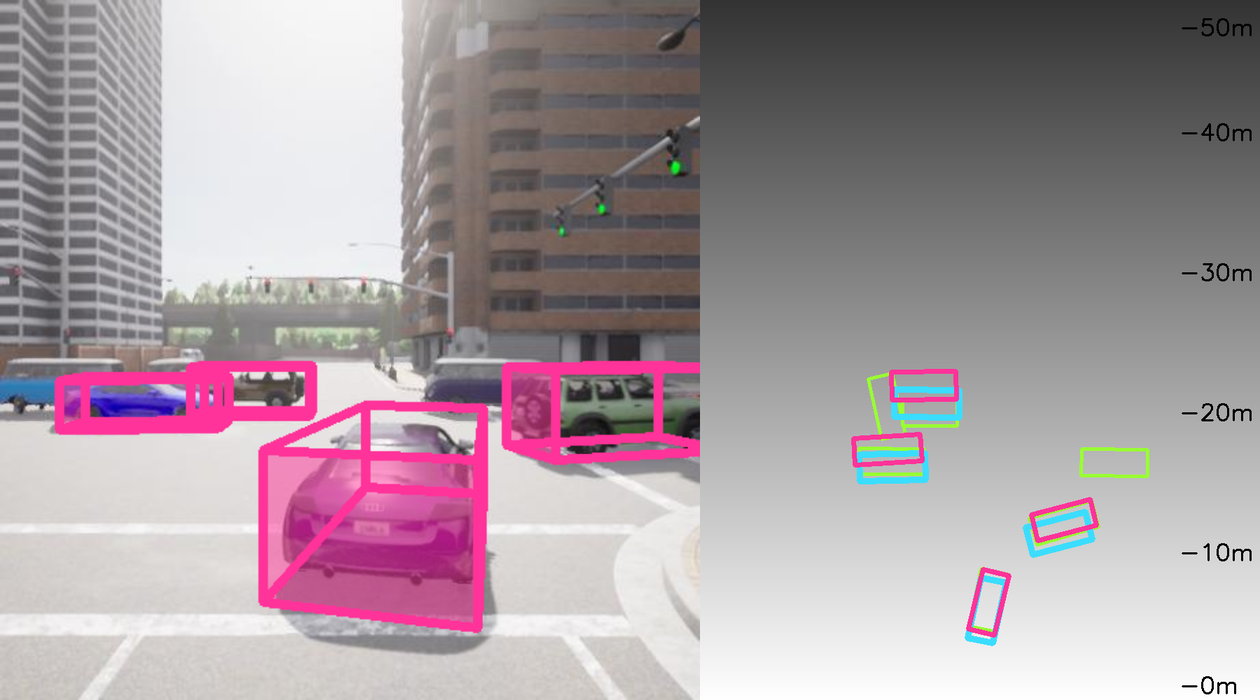}
            \end{subfigure}
            \begin{subfigure}{0.4\linewidth}
                \includegraphics[width=\linewidth]{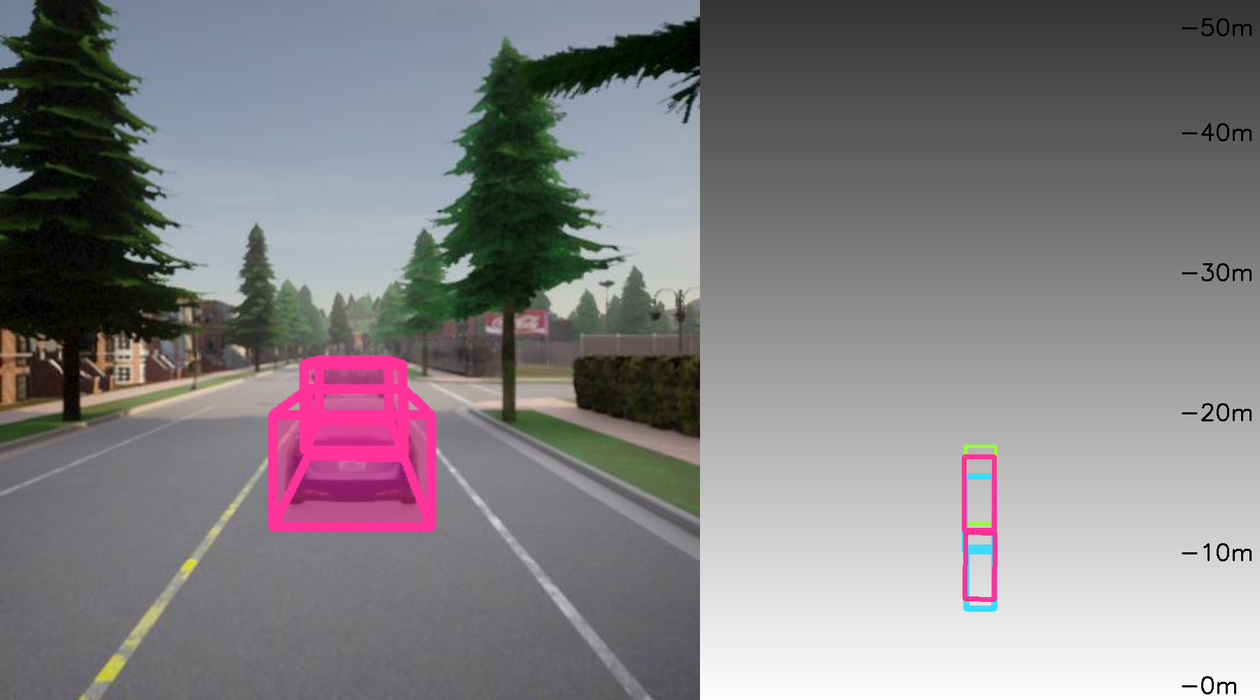}
            \end{subfigure}
            \caption{\textbf{\carla \val Qualitative Results}.
            \methodName detects objects more accurately than \gupNet\cite{lu2021geometry}, making \methodName more robust to camera height changes.
            The \textcolor{cyan}{regression-based baseline \gupNet} mostly underestimates the depth which qualitatively justifies the claims of \cref{theorem:2}.
            All methods are trained on \carla images at car height $\egoHeightChange=0m$ and evaluated on $\egoHeightChange=+0.76m$.
            [Key: \textcolor{my_magenta}{Cars} of \methodName.
            ; \textcolor{cyan}{Cars} of \gupNet, and \textcolor{green}{Ground Truth} in BEV.
            }
            \label{fig:qualitative_carla}
        \end{figure*}

            \noIndentHeading{\carla.}
                We now show some qualitative results of models trained on \carla \val split from car height $(\egoHeightChange=0m)$ and tested on truck height $(\egoHeightChange=+0.76m)$ in \cref{fig:qualitative_carla}.
                We depict the predictions of \methodName in image view on the left, the predictions of \methodName, the baseline \gupNet \cite{lu2021geometry}, and GT boxes in \bev on the right.
                In general, \methodName detects objects more accurately than \gupNet\cite{lu2021geometry}, making \methodName more robust to camera height changes.
                The regression-based baseline \gupNet mostly underestimates the depth of \threeD boxes with positive ego height changes, which qualitatively justifies the claims of \cref{theorem:2}.

\begin{figure*}[!t]
            \centering
            \begin{subfigure}{0.48\linewidth}
                \includegraphics[width=\linewidth]{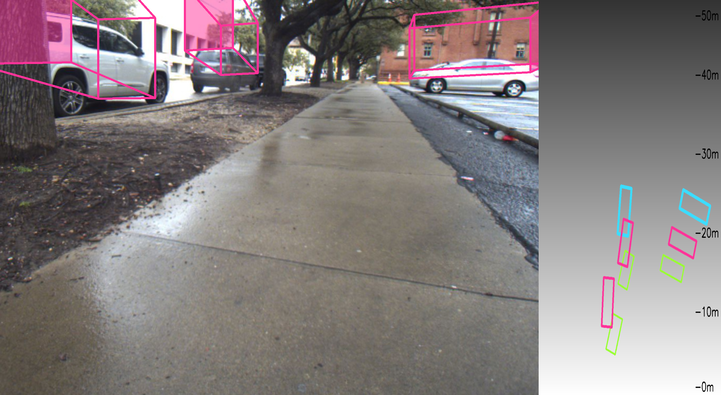}
            \end{subfigure}
            \begin{subfigure}{0.48\linewidth}
                \includegraphics[width=\linewidth]{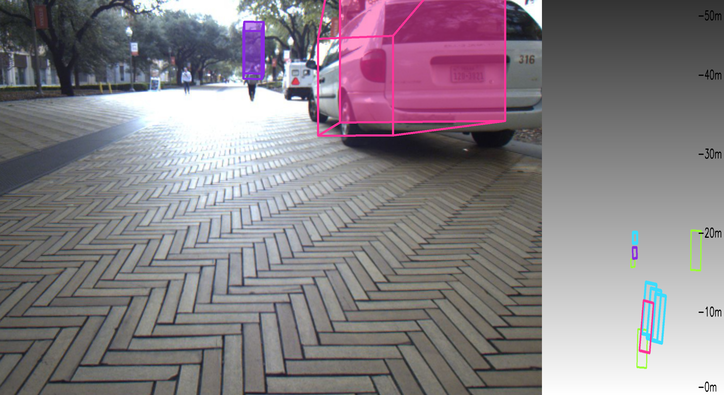}
            \end{subfigure}
            \begin{subfigure}{0.48\linewidth}
                \includegraphics[width=\linewidth]{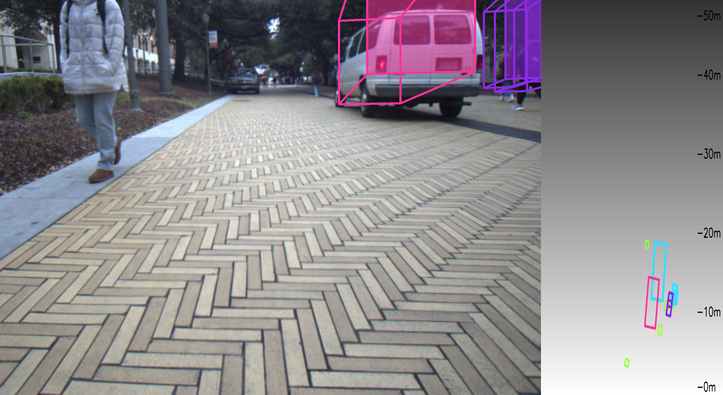}
            \end{subfigure}
            \begin{subfigure}{0.48\linewidth}
                \includegraphics[width=\linewidth]{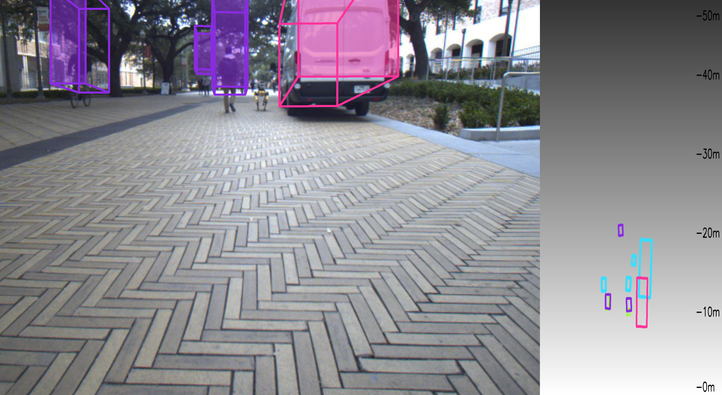}
            \end{subfigure}
            \begin{subfigure}{0.48\linewidth}
                \includegraphics[width=\linewidth]{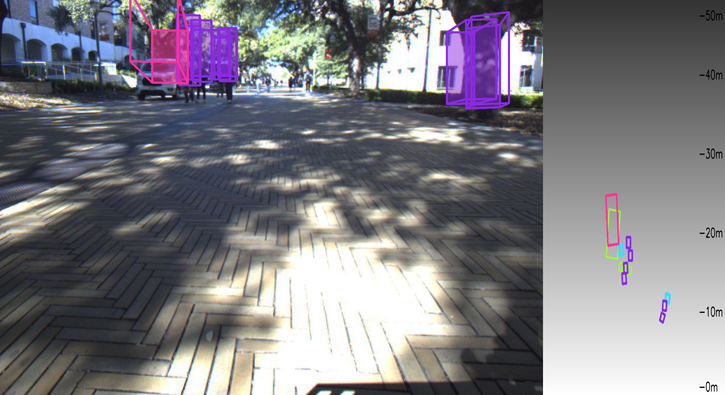}
            \end{subfigure}
            \begin{subfigure}{0.48\linewidth}
                \includegraphics[width=\linewidth]{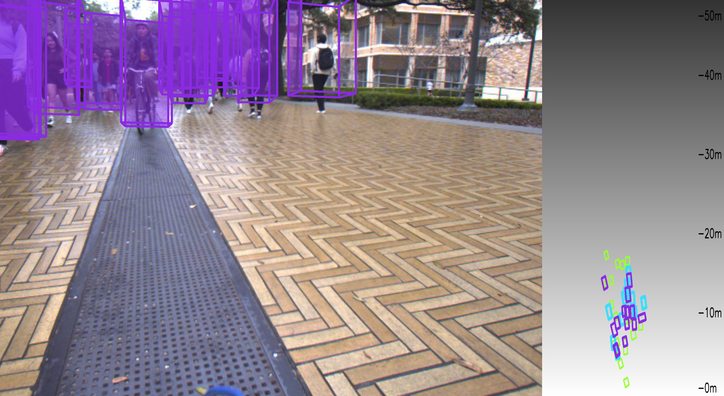}
            \end{subfigure}
            \begin{subfigure}{0.48\linewidth}
                \includegraphics[width=\linewidth]{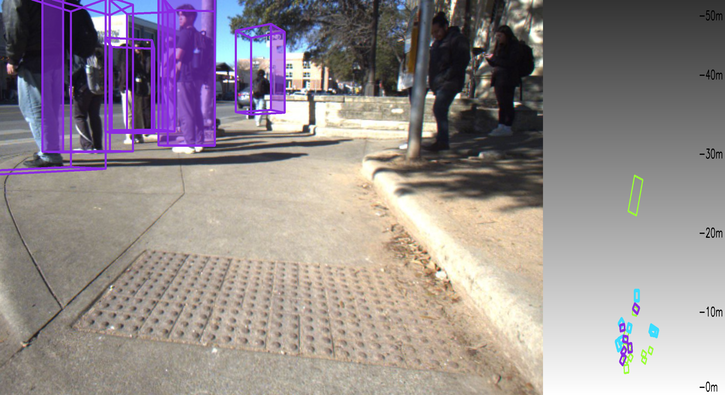}
            \end{subfigure}
            \begin{subfigure}{0.48\linewidth}
                \includegraphics[width=\linewidth]{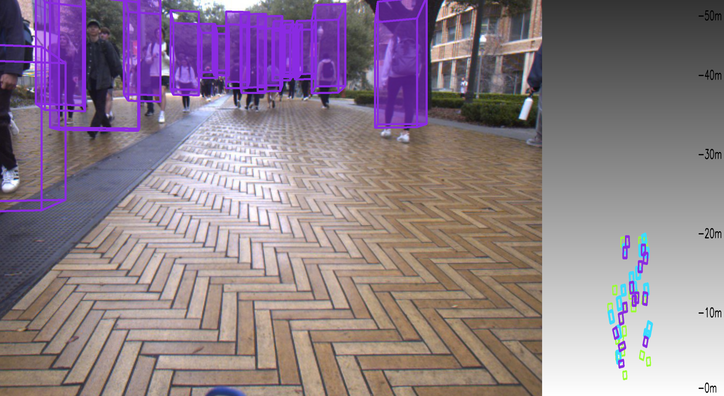}
            \end{subfigure}
            \caption{\textbf{\coda \val Qualitative Results}.
            \methodName detects objects more accurately than \gupNet\cite{lu2021geometry}, making \methodName more robust to camera height changes.
            All methods are trained on \nuscenes dataset and evaluated on \coda dataset.
            [Key: \textcolor{my_magenta}{Cars} and \textcolor{violet}{Pedestrian} of \methodName.
            ; \textcolor{cyan}{all classes} of \gupNet, and \textcolor{green}{Ground Truth} in BEV.
            }
            \label{fig:qualitative_coda}
        \end{figure*}

            \noIndentHeading{\coda.}
                We now show some qualitative results of models trained on \coda \val split in \cref{fig:qualitative_coda}.
                As before, we depict the predictions of \methodName in image view image view on the left, the predictions of \methodName, the baseline \gupNet \cite{lu2021geometry}, and GT boxes in \bev on the right.
                In general, \methodName detects objects more accurately than the baseline \gupNet\cite{lu2021geometry}, making \methodName more robust to camera height changes.
                Also, considerably less number of boxes are detected in the cross-dataset evaluation \thatIs{} on \coda \val.
                We believe this happens because of the domain shift.

\end{document}